\documentclass{article}

\usepackage[nonatbib, preprint]{neurips_2022}

\usepackage{microtype}
\usepackage{graphicx}
\usepackage{subfigure}
\usepackage{booktabs} 
\usepackage{algorithm2e}
\RestyleAlgo{ruled}
\usepackage{multirow}
\usepackage{enumitem}
\usepackage{xcolor}
\usepackage{hyperref}

\usepackage[utf8]{inputenc} 
\usepackage[T1]{fontenc}    
\usepackage{url}            
\usepackage{booktabs}       
\usepackage{amsfonts}       
\usepackage{nicefrac}       
\usepackage{microtype}      
\usepackage{xcolor}         
\usepackage{adjustbox}
\input{mysymbol.sty}
\usepackage{balance}
\usepackage{graphicx}
\usepackage[utf8]{inputenc} 
\usepackage[T1]{fontenc}    
\usepackage{url}            
\usepackage{booktabs}       
\usepackage{amsfonts}       
\usepackage{nicefrac}       
\usepackage{microtype}      
\usepackage{xcolor}         
\usepackage{multicol}
\usepackage{amssymb,amsmath,amsthm, amsfonts}
\newtheorem{theorem}{Theorem}
\newtheorem{lemma}{Lemma}

\definecolor{morange}{rgb}{0.8,0.2,0}
\definecolor{mblue}{rgb}{0,0.3,1.0}
\definecolor{mred}{rgb}{0.9,0.1,0.1}
\definecolor{mpurple}{rgb}{0 0 0}

\allowdisplaybreaks[4]

\title{FairGAT: Fairness-aware Graph Attention Networks
}

\author{
  O. Deniz Kose  \\
  Department of Electrical Engineering and Computer Science\\
  University of California, Irvine\\
  \texttt{okose@uci.edu} \\
\And
  Yanning Shen \\
  Department of Electrical Engineering and Computer Science\\
  University of California, Irvine\\
  \texttt{yannings@uci.edu} \\
}

\begin{document}

\maketitle

\begin{abstract}
Graphs can facilitate modeling various complex systems such as gene networks and power grids, as well as analyzing the underlying relations within them. Learning over graphs has recently attracted increasing attention, particularly graph neural network-based (GNN) solutions, among which graph attention networks (GATs) have become one of the most widely utilized neural network structures for graph-based tasks. Although it is shown that the use of graph structures in learning results in the amplification of algorithmic bias, the influence of the attention design in GATs on algorithmic bias has not been investigated. Motivated by this, the present study first carries out a theoretical analysis in order to demonstrate the sources of algorithmic bias in GAT-based learning for node classification. Then, a novel algorithm, FairGAT, that leverages a fairness-aware attention design is developed based on the theoretical findings. Experimental results on real-world networks demonstrate that FairGAT improves group fairness measures while also providing comparable utility to the fairness-aware baselines for node classification and link prediction.

\end{abstract}
\section{Introduction}
\par We live in the era of connectivity, where the behaviors of humans and devices are increasingly driven by their relations to others. Thus, a significant amount of data is collected from various interconnected systems, such as social networks \cite{social}, power grid networks \cite{grid}, and gene networks \cite{gene} to name a few. Learning from such data can hence benefit better understanding and designing the corresponding networked systems. This motivates the increasing attention towards learning over graphs \cite{chami2022machine}. Among different approaches, graph neural networks (GNNs) have become the state-of-the-art in graph-based learning due to their success in various tasks \cite{matrix,soc,chem}.

GNNs employ different aggregation mechanisms to obtain node embeddings by aggregating the representations of neighbors. Among different GNN layers, graph attention networks (GATs) \cite{gat} have become one of the most widely utilized GNN designs \cite{zhou2020graph}. GATs improve the conventional aggregation schemes over graph structure by leveraging masked self-attention layers. Specifically, instead of assigning uniform weights to each neighbor hence assuming each neighbor to be of the same importance, the attention mechanism learns neighbor-specific weights. Such non-uniform weights result in a more flexible aggregation framework, and can find the more relevant neighbors. 

 Algorithmic bias refers to the performance gap incurred by machine learning (ML) algorithms with respect to certain sensitive/protected attributes (e.g., gender, ethnicity) in the context of group fairness \cite{pedreshi2008discrimination}. For example, the accuracy differences between groups of people from different ethnicities in a face recognition model correspond to the algorithmic bias with respect to the sensitive attribute race. 
 It has been demonstrated that 
 GNN-based frameworks not only propagate but may also amplify the already existing algorithmic bias due to the utilization of 
biased structural information \cite{say}.  
For example, nodes in social networks tend to connect to other nodes with similar attributes, which leads to a denser connectivity between the nodes with the same sensitive attributes \cite{segregation}. Hence, by aggregating information from the neighbors, the representations obtained by GNNs may be highly correlated with the sensitive attributes. This causes indirect discrimination in the ensuing learning tasks, even when the sensitive attributes are not directly used in training \cite{indirect}. 

Despite the increasing popularity of GATs, the impact of attention on algorithmic bias has not been studied to the best of our knowledge. The present work analyzes the sources of bias in a GAT-based neural network trained for node classification. Specifically, our analysis characterizes all the factors that play a role in the parity of predictions for different sensitive groups. Based on the theoretical findings for the sources of bias, a novel algorithm, called FairGAT, is designed. FairGAT introduces a fairness-aware attention layer that can mitigate bias via a novel attention learning strategy.
The proposed algorithm improves fairness while providing comparable utility to state-of-the-art fairness-aware schemes, is efficient in terms of time complexity, and can be flexibly employed together with other fairness enhancement methods. The
contributions of this work can be summarized as follows:\\
    \textbf{c1)} For a neural network consisting of multiple GAT layers, we present a theoretical analysis that illuminates the sources of bias leading to disparity between the predictions for different sensitive groups.\\
  \textbf{c2)} Presented analysis strategy can pave the way for further theoretical findings illustrating the sources of bias for different GNN layers, e.g., graph convolutional networks \cite{supervised1}. \\
    \textbf{c3)} Based on the developed analysis, we devise a novel algorithm, FairGAT, with three main steps. Each step in the algorithm combats one of the identified sources of bias for a GAT-based neural network. \\
    \textbf{c4)} FairGAT introduces a new attention mechanism that can mitigate algorithmic bias. The proposed fair attention learning strategy is efficient, i.e., it does not incur significant additional computational complexity compared to the conventional GATs. \\
    \textbf{c5)} The experimental results are obtained over six real-world networks for node classification and link prediction. The comparative results show that FairGAT can improve group fairness measures while providing similar utility to the fairness-aware baselines.  \\

\section{Related Work}

\noindent\textbf{ML over graphs.} Conventional graph learning approaches can be summarized under two categories: factorization-based and random walk-based approaches. Factorization-based methods minimize the difference between the inner product of the created node representations and a deterministic similarity metric (that is typically based on the graph structure) between the nodes, e.g., \cite{factorization1, factorization2, factorization3}. Random walk-based approaches, on the other hand, employ stochastic measures of similarity and target at node representations whose inner products reflect the considered stochastic similarity measure between the nodes, e.g., \cite{deepwalk, node2vec, line, harp}.
Recently, GNNs have gained popularity and have become the state-of-the-art for a number graph-based tasks, see e.g., \cite{supervised3,supervised4,unsup,graphsage, grace, adaptive, dgi}. 

\textbf{Fairness-aware learning over graphs.} \cite{fairwalk} serves as a seminal work for fairness-aware random walk-based studies \cite{khajehnejad2022crosswalk}. In addition, \cite{say,compositional,debiasing} propose to use adversarial regularizers to mitigate bias in GNNs. Another strategy is to utilize a Bayesian approach where the sensitive information is modeled in the prior distribution to enhance fairness over graphs \cite{debayes}. Furthermore, \cite{subgroup} performs a PAC-Bayesian analysis and links the notion of subgroup generalization to accuracy disparity, and \cite{heterogeneous} proposes several strategies, including GNN-based ones, to reduce bias for the representations of heterogeneous information networks. Specifically for fairness-aware link prediction, \cite{kl} introduces a regularizer, \cite{dyadic,all} propose strategies that alter the adjacency matrix. With a specific consideration of individual fairness over graphs, \cite{individual} proposes a ranking-based framework. An alternative approach in fairness-aware graph-based learning is to modify the graph structure and nodal features to combat bias via automated or manually designed fair graph data augmentations \cite{nifty,biased,icml,tsipn, dong2022edits}. Differing from the majority of these prior works, our work first develops a theoretical analysis for algorithmic bias for our system of interest (GATs), based on which it proposes a systematic framework to mitigate the bias.
\begin{table}[t]
    \centering
    \begin{footnotesize}
    \begin{tabular}{ c | c }
         \toprule
  \textbf{Notations} & \textbf{Definitions or Descriptions }\\
\toprule
     { $\mathbf{X}$}   &      { nodal features}\\
   {   $\mathbf{A}$}  & { adjacency matrix}\\  
    {  $\mathbf{s}$}  & { sensitive attribute vector}\\
    {  $\mathbf{H}^{l+1}$}  & {node representations created at layer $l$}\\ 
    { $\mathcal{S}_{i}$}  & {set of nodes with sensitive attribute $i$}\\ 
   {  $\mathcal{E}^{\chi}$} & {set of inter-edges} \\ 
     {$\mathcal{E}^{\omega}$ }& {set of intra-edges} \\ 
    { $ \mathcal{S}^{\chi}$} & {set of nodes with at least one inter-edge}\\ 
    { $\mathcal{S}_{i}^{\chi}$ }& {set of nodes in the intersection of $\mathcal{S}_{i}$ and $\mathcal{S}^{\chi}$}\\ 
    { $ \mathcal{S}^{\omega}$} & { set of nodes with no inter-edges}\\ 
    { $\mathcal{S}_{i}^{\omega}$} & {set of nodes in the intersection of $\mathcal{S}_{i}$ and $\mathcal{S}^{\omega}$}\\ 
   \bottomrule
    \end{tabular}
    \end{footnotesize}
    \caption{List of symbols.}
    \label{table:notation}
\end{table}
\section{Preliminaries}
\par The focus of this study is designing a fair GAT-based neural network for node classification for a given graph $\mathcal{G}:=(\mathcal{V}, \mathcal{E})$, where $\mathcal{V}:=$ $\left\{v_{1}, v_{2}, \cdots, v_{N}\right\}$ is the set of nodes and $\mathcal{E} \subseteq \mathcal{V} \times \mathcal{V}$ denotes the set of edges. Nodal features and graph adjacency matrices of the input graph $\mathcal{G}$ are denoted by $\mathbf{X} \in \mathbb{R}^{N \times F}$ and $\mathbf{A} \in\{0,1\}^{N \times N}$, respectively, where $\mathbf{A}_{i j}=1$ if and only if $\left(v_{i}, v_{j}\right) \in \mathcal{E}$. This work considers a single, binary sensitive attribute for each node which is denoted by $\mathbf{s} \in \{0,1\}^{N}$. The learned node representations in the neural network at layer $l$ are denoted by $\mathbf{H}^{l+1}$. $\mathbf{x}_{i} \in \mathbb{R}^{F}$, $\mathbf{h}^{l+1}_{i} \in \mathbb{R}^{F^{l}}$, and $s_{i} \in \{0,1\}$ denote the feature vector, representation at layer $l$ and the sensitive attribute of node $v_{i}$, respectively. Furthermore, $\mathcal{S}_{0}$ and $\mathcal{S}_{1}$ denote the set of nodes whose sensitive attributes are $0$ and $1$, respectively. Define inter-edge set $\mathcal{E}^{\chi}:=\{e_{ij}|v_i \in \mathcal{S}_a, v_j \in \mathcal{S}_b, a\neq b\}$, while intra-edge set is defined as  $\mathcal{E}^{\omega}:= \{e_{ij}|v_i \in \mathcal{S}_a, v_j \in \mathcal{S}_b, a= b\}$. Similarly, the set of nodes having at least one inter-edge is denoted by $\mathcal{S}^{\chi}$, while $\mathcal{S}^{\omega}$ defines the set of nodes that have no inter-edges (i.e., only have intra-edges).  The intersection of the sets $\mathcal{S}_{0}$ and $\mathcal{S}^{\chi}$ is denoted by $\mathcal{S}_{0}^{\chi}$. 
Table \ref{table:notation} summarizes the overall notation.


GNNs create node embeddings by aggregating the representations of neighbors for each node. Most GNN structures assign equal weights to all neighbors in the information aggregation with an implicit assumption that all neighbors have the same importance to the anchor node. On the other hand, GATs learn weights $\alpha_{ij}$ which indicates the importance of neighbor node $j$ to the anchor node $i$. Via learning attention coefficients $\alpha_{ij}$, GATs can select the most relevant neighbors to the anchor node, which results in a more flexible framework than equal weight assignment. 

The attention learning and information a ggregation process in a conventional $l$th GAT layer can be summarized as follows \cite{gat}:
\begin{enumerate}
    \item  $e\left(\boldsymbol{h}_i^{l}, \boldsymbol{h}^{l}_j\right) := \operatorname{LReLU}\left((\boldsymbol{a}^{l})^{\top} \cdot\left[\boldsymbol{W}^{l} \boldsymbol{h}^{l}_i \| \boldsymbol{W} ^{l}\boldsymbol{h}^{l}_j\right]\right)$,
\item $\alpha^{l}_{i j}=\frac{\exp \left(e\left(\boldsymbol{h}^{l}_i, \boldsymbol{h}^{l}_j\right)\right)}{\sum_{j^{\prime} \in \mathcal{N}_i} \exp \left(e\left(\boldsymbol{h}^{l}_i, \boldsymbol{h}^{l}_{j^{\prime}}\right)\right)}$,
\item $\boldsymbol{h}_i^{l+1}=\sigma\left(\sum_{j \in \mathcal{N}_i} \alpha^{l}_{i j} \cdot \boldsymbol{W}^{l} \boldsymbol{h}^{l}_j\right)$.
\end{enumerate}
Here, $\mathbf{a}^{l}$, $\mathbf{W}^{l}$ are learnable parameters of the GAT at layer $l$. Furthermore, $\mathcal{N}_i$ denotes the set of neighbors of node $v_{i}$, while LReLU stands for LeakyReLU \cite{leakyrelu}, and $\sigma(\cdot)$ is the utilized non-linear activation. 

\section{Methodology}
\label{sec:fairgat}
This section first investigates the sources of bias in a GAT-based neural network that is trained for node classification. 
As the bias measure, the disparity between the predictions for different sensitive groups are utilized:
\begin{equation}
     \delta_{\hat{y}}:=\left\|\operatorname{mean}(\mathbf{\hat{y}}_{j} \mid s_{j}=0) - \operatorname{mean}(\mathbf{\hat{y}}_{j} \mid s_{j}=1) \right\|_{2},
\end{equation}
{\color{mpurple}where $\mathbf{\hat{y}}_{j}$ denotes the predicted soft label for node $v_{j}$ and $\operatorname{mean}(\cdot)$ is the sample mean operator. Note that $\delta_{\hat{y}}$ generalizes the commonly utilized group fairness metric statistical parity, $\Delta_{S P}:=|P(\hat{c}_{j}=1 \mid s_{j}=0)-P(\hat{c}_{j}=1 \mid s_{j}=1)|$ where $\hat{c}$ stands for the prediction of hard/class labels, when $\mathbf{\hat{y}}_{j}$ is output by the sigmoid activation and denotes the probability that $v_{j}$ has the class label of $1$. Specifically, focusing on the term $P(\hat{c}_{j}=1 \mid s_{j}=0)$, it follows that:
\begin{equation}
\begin{split}
    P(\hat{c}_{j}=1& \mid s_{j}=0) 
    \\&= \int_{0}^{1} P(\hat{c}_{j}=1 \mid \hat{y}_{j}, s_{j}=0) P(\hat{y}_{j} \mid s_{j}=0) d\hat{y}_{j} \\
    & = \int_{0}^{1} P(\hat{c}_{j}=1 \mid \hat{y}_{j}) P(\hat{y}_{j} \mid s_{j}=0) d\hat{y}_{j}, 
    \end{split}
\end{equation}
where the last equality follows from the fact that the random variables $s_{j} \rightarrow \hat{y}_{j} \rightarrow \hat{c}_{j}$ form a Markov chain. Furthermore, $P(\hat{c}_{j}=1 \mid \hat{y}_{j}) = \hat{y}_{j}$, as $\hat{y}_{j}$ is assumed to be a soft label denoting the probability that $v_{j}$ has the class label of $1$, leading to:
\begin{equation}
\begin{split}
    P(\hat{c}_{j}=1& \mid s_{j}=0) = \int_{0}^{1} \hat{y}_{j} P(\hat{y}_{j} \mid s_{j}=0) d\hat{y}_{j}, \\
    &=\operatorname{mean}(\hat{y}_{j} \mid s_{j}=0).
\end{split}
\end{equation}
The same can be derived for $ P(\hat{c}_{j}=1 \mid s_{j}=1)$, where $ P(\hat{c}_{j}=1 \mid s_{j}=1) = \operatorname{mean}(\hat{y}_{j} \mid s_{j}=1)$, proving that $\Delta_{S P}$ is a special case of $\delta_{\hat{y}}$ when $\mathbf{\hat{y}}_{j}$  is output by sigmoid activation and denotes the probability that $v_{j}$ has the class label of $1$.}
After the sources of bias are demonstrated, our proposed algorithm, FairGAT, is developed and presented. 
\subsection{Bias Analysis}
\label{subsec:theory}
This subsection aims to illuminate the factors that lead to the disparity between the predictions for different sensitive groups, $\delta_{\hat{y}}$, in a GAT-based network trained for node classification. Let $\mathbf{Z}^{l+1}$ denote the aggregated representations by the $l$th GAT layer with $i$th row $\mathbf{z}^{l+1}_{i}:= \sum_{j \in \mathcal{N}_{i}} \alpha^{l}_{i j} \mathbf{c}^{l+1}_{j}$, where $\mathbf{c}^{l+1}_{i}:=\mathbf{W}^{l} \mathbf{h}^{l}_{i}$. The sample means of  $\mathbf{c}^{l+1}$ and $\mathbf{z}^{l+1}$ vectors are represented by 
$\bar{\mathbf{c}}_{s}^{l+1} :={\rm mean}(\mathbf{c}^{l+1}_j \mid v_{j} \in \mathcal{S}_{s})$ and $
\bar{\mathbf{z}}_{s}^{l+1} :={\rm mean}(\mathbf{z}^{l+1}_j \mid v_{j} \in \mathcal{S}_{s})$ for the nodes in sensitive group $\mathcal{S}_{s}$ for $s=0,1$. 
The following assumptions are made for the theoretical findings in this work:
\\ \textbf{A1: }$\left\|\mathbf{c}^{l+1}_j-\bar{\mathbf{c}}_{s}^{l+1}\right\|_{\infty} \leq (\Delta_{c}^{(s)})^{l+1}$, $\forall v_{j} \in \mathcal{S}_{s}$ with $s \in \{0,1\}$, where $\Delta^{l+1}_{c} = \operatorname{max}((\Delta_{c}^{(0)})^{l+1}, (\Delta_{c}^{(1)})^{l+1})$. $\left\|\mathbf{z}^{l+1}_j-\bar{\mathbf{z}}_{s}^{l+1}\right\|_{\infty} \leq (\Delta_{z}^{(s)})^{l+1}$, $\forall v_{j} \in \mathcal{S}_{s}$ with $s \in \{0,1\}$, where $\Delta^{l+1}_{z} = \operatorname{max}((\Delta_{z}^{(0)})^{l+1}, (\Delta_{z}^{(1)})^{l+1})$. Here, $\operatorname{max}(\cdot,\cdot)$ outputs the element-wise maximum of the input vectors. 
\\ { \textbf{A2: } The total amount of attention that is assigned to the neighbors from different sensitive groups is the same for every node $v_{k} \in \mathcal{V}$ at every layer $l$, where $\alpha^{\chi} = \alpha_{k}^{\chi}=\sum_{a \in \mathcal{N}(k) \cap S_i} \alpha_{k a},  \forall v_k \in \mathcal{S}_j \text{ if  } i \neq j$.} Consequently, the total amount of attention that is assigned to the neighbors from the same sensitive group is also the same for every node $v_{k} \in \mathcal{V}$ at every layer $l$, i.e., $\alpha^{\omega} = \alpha_{k}^{\omega}=\sum_{a \in \mathcal{N}(k) \cap S_i} \alpha_{k a},  \forall v_k \in \mathcal{S}_j \text{ if  } i = j$.)

Based on these assumptions, Theorem \ref{theorem:gat} demonstrates the factors that contribute to the disparity between the representations of different sensitive groups obtained at the $l$th GAT layer. Specifically, Theorem \ref{theorem:gat} upper bounds the term $\delta_{h}^{l+1}:=\left\|\operatorname{mean}(\mathbf{h}^{l+1}_{j} \mid s_{j}=0) - \operatorname{mean}(\mathbf{h}^{l+1}_{j} \mid s_{j}=1)\right\|_{2}$. The proof of the theorem is presented in Appendix \ref{app:theorem_proof}.

\begin{theorem}
    \label{theorem:gat}
    The disparity between the representations of different sensitive groups that are output by the $l$th GAT layer, $\delta_{h}^{l+1}$, can be upper bounded by:
    \begin{equation}\label{delta_h1}
       \delta_{h}^{l+1} \leq L \Big(  \sigma_{max}(\mathbf{W}^{l}) \big| (R_{1}^{\chi}\alpha^{\chi} + R_{0}^{\chi}\alpha^{\chi} - 1)\big| \delta_{h}^{l} + 2 \sqrt{N} \Delta^{l+1}_{c} + 2 \sqrt{N} \Delta^{l+1}_{z} \Big), 
    \end{equation}
    where 
    $L$ is the Lipschitz constant of nonlinear activation $\sigma$, $ \sigma_{max}(\cdot)$ denotes the largest singular value of the input matrix, and  $R_{1}^{\chi}:=\frac{|S^{\chi}_{1}|}{|\mathcal{S}_{1}|}, R_{0}^{\chi}:=\frac{|S^{\chi}_{0}|}{|\mathcal{S}_{0}|}$.
\end{theorem}

Theorem \ref{theorem:gat} explains the factors that can amplify the disparity of representations throughout a GAT layer. We note that in the majority of our baselines herein, we observed that a single, final, fully connected layer is employed for supervised node classification, after multiple GNN layers. Thus, we expand our analysis in Theorem \ref{theorem:gat} to systems that also include a fully connected layer, through Lemma \ref{lemma:mlp}. Specifically, Lemma \ref{lemma:mlp} investigates the sources of bias by upper bounding $\delta_{h}^{l+1}$ for a fully connected layer, whose proof is provided in Appendix \ref{app:lemma_mlp}. 
\begin{lemma}
\label{lemma:mlp}
The disparity between the representations from different sensitive groups that are output by the $l$th fully connected layer with input-output relationship $\mathbf{H}^{l+1} = \sigma(\mathbf{H}^{l} \mathbf{W}^{l})$, $\delta_{h}^{l+1}$, can be upper bounded by:
    \begin{equation}\label{delta_h2}
       \delta_{h}^{l+1} \leq L (  \sigma_{max}(\mathbf{W}^{l}) \delta_{h}^{l} + 2 \sqrt{N} \Delta^{l+1}_{z}),
    \end{equation}
    where 
    $\mathbf{W}^{l}$ is the learnable weight matrix at the $l$th fully connected layer and $\mathbf{Z}^{l+1} = \mathbf{H}^{l} \mathbf{W}^{l}$.
\end{lemma}

\textbf{Remark. }The results of Theorem \ref{theorem:gat} together with Lemma \ref{lemma:mlp} can help characterize the sources of bias in a neural network with multiple GAT layers, possibly followed by a fully connected layer. To exemplify a special case, consider a neural network consisting of a single GAT layer with input-output relation $\boldsymbol{h}_i^{1}=\sigma\left(\sum_{j \in \mathcal{N}_i} \alpha^{0}_{i j} \cdot \boldsymbol{W}^{0} \boldsymbol{x}^{0}_j\right)$, followed by a fully connected layer with input-output relation $\mathbf{\hat{y}}_{i}= \mathbf{h}^{2}_{i} =\sigma\left(\boldsymbol{W}^{1} \mathbf{h}^{1}_i\right)$. Lemma \ref{lemma:mlp} implies that the disparity between the outputs of different sensitive groups, $\delta_{\hat{y}}:=\left\|\operatorname{mean}(\mathbf{\hat{y}}_{j} \mid s_{j}=0) - \operatorname{mean}(\mathbf{\hat{y}}_{j} \mid s_{j}=1) \right\|_{2}$, can be upper bounded by:
    \begin{equation}
     \label{eq:main_res1}
         \delta_{\hat{y}} \leq L (  \sigma_{max}(\mathbf{W}^{1}) \delta_{h}^{1} + 2 \sqrt{N} \Delta^{2}_{z}),
    \end{equation}
    and Theorem \ref{theorem:gat} further shows that
    \begin{equation}
    \label{eq:main_res2}
     \delta_{h}^{1} \leq  L (  \sigma_{max}(\mathbf{W}^{0}) |(R_{1}^{\chi}\alpha^{\chi} + R_{0}^{\chi}\alpha^{\chi} - 1)| \delta_{x} + 2 \sqrt{N} \Delta^{1}_{c} + 2 \sqrt{N} \Delta^{1}_{z}), 
    \end{equation}
    where $\delta_{x}:=\left\|\operatorname{mean}(\mathbf{x}_{j} \mid s_{j}=0) - \operatorname{mean}(\mathbf{x}_{j} \mid s_{j}=1) \right\|_{2}$.
 The bias measure $\delta_{\hat{y}}$ for the overall network can be upper bounded based on \eqref{eq:main_res1} and \eqref{eq:main_res2}. Our framework, FairGAT, motivates by lowering this upper bound so that the overall scheme leads to a smaller $\delta_{\hat{y}}$ value. Note that while the results in \eqref{eq:main_res1} and \eqref{eq:main_res2} are presented for a network with a single GAT layer, they can easily be extended to multiple layers of GATs by utilizing Theorem \ref{theorem:gat} and Lemma \ref{lemma:mlp}. 

Overall, our analysis illuminates the factors that play a role in the propagated bias towards the predictions of a GAT-based neural network, which also hints at the possible solutions to combat such bias. Since $\delta_{h}^{l}$  characterizes the disparity between the output representations at layer $l$, the relevant terms in \eqref{delta_h1} and \eqref{delta_h2} should be properly controlled to avoid the amplification of bias. Specifically, the following steps can be applied in order to design a fair GAT-based network:
\begin{enumerate}
    \item \textbf{Fair attention learning: }Theorem \ref{theorem:gat} implies that the total amount of attention assigned to inter-edges, $\alpha^{\chi}$, can be manipulated to reduce the resulting intrinsic bias, since $\delta_{\hat{y}}$ is a function of $|(R_{1}^{\chi}\alpha^{\chi} + R_{0}^{\chi}\alpha^{\chi} - 1)|$.
    
    \item \textbf{Spectral normalization of weight matrices: }The analysis shows that the propagated bias towards the predictions is affected by the spectral properties of the weight matrices $\mathbf{W}^{l}$ at every layer $l$. Specifically, it can be observed from the upper bound of $\delta_{h}^{l}$ in \eqref{delta_h1} and \eqref{delta_h2} that
    the largest singular value $\sigma_{max}(\mathbf{W}^{l})$ (i.e., spectral norm) of the weight matrix should not be larger than $1$ in order not to amplify the already existing disparity. Therefore, spectral normalization is applied to $\mathbf{W}^{l}$ at every layer $l$ to guarantee that $\sigma_{max}(\mathbf{W}^{l}) \leq 1$.  
    
    \item \textbf{Scaling representations: }Finally, Theorem \ref{theorem:gat} and Lemma \ref{lemma:mlp} suggest that the maximal deviation $\Delta_{c}$ at every attention layer, and $\Delta_{z}$ at every attention or fully connected layer, influence the bias measure $\delta_{\hat{y}}$. Such deviations can be {manipulated for bias mitigation.}
\end{enumerate}
\textbf{Remark. }Theorem \ref{theorem:gat} also suggests that the disparity between the nodal features of different sensitive groups, $\delta_{x}$, should be decreased for a lower upper bound on $\delta_{\hat{y}}$. Since $\delta_{x}$ is not influenced by the changes in other factors, the present framework will not focus on this term. However, the proposed framework herein can be employed in conjunction with a fairness-aware nodal feature manipulation strategy, such as the ones developed in \cite{kose2022fair2, dong2022edits}, to reduce $\delta_{x}$.

\subsection{Proposed Scheme: FairGAT}
\label{subsec:meth}
Building upon the theoretical analysis in Subsection \ref{subsec:theory}, the present subsection develops a novel framework to reduce $\delta_{\hat{y}}$ for a GAT-based network trained for node classification. The overall scheme includes three steps as mentioned before, the design of which will be discussed in detail. The overall algorithm is presented in Algorithm \ref{alg:fairgat}.


\subsubsection{Fair attention learning } Theorem \ref{theorem:gat} demonstrates that the total amount of attention assigned to inter-edges, $\alpha^{\chi}$, can be manipulated to lower the term $|(R_{1}^{\chi}\alpha^{\chi} + R_{0}^{\chi}\alpha^{\chi} - 1)|$ for bias mitigation. Motivated by this, a fairness-aware graph-attention layer is designed in this step, where the learned attention minimizes the bias-related term $|(R_{1}^{\chi}\alpha^{\chi} + R_{0}^{\chi}\alpha^{\chi} - 1)|$. To this end, the optimal amount of attention that should be assigned to the inter-edges is analyzed herein. Specifically, we consider the following optimization problem: 
\begin{equation}
\label{eq:optim}
\begin{array}{rrclcl}
\displaystyle (\alpha^{\chi})^{*} = \min_{\alpha^{\chi}} & \multicolumn{3}{l}{|R_{1}^{\chi}\alpha^{\chi} + R_{0}^{\chi}\alpha^{\chi} - 1|}\\
\textrm{s.t.} & 0 \leq \alpha^{\chi} \leq \alpha^{\chi}_{max}. \\
\end{array}
\end{equation}
Note that $\alpha^{\chi} + \alpha^{\omega} =1$ due to the utilized softmax activation in the calculation of attention in the conventional GAT layers. 
Therefore, the optimal amount of attention that should be assigned to the intra-edges equals to $(\alpha^{\omega})^{*} = 1 - (\alpha^{\chi})^{*}$. In equation \ref{eq:optim}, $\alpha^{\chi}_{max} \leq 1$ specifies the maximum amount of attention that can be assigned to the inter-edges, thus is a hyperparameter used to provide a trade-off between the utility and fairness. Having $(\alpha^{\chi})^{*}=1$ would mean $(\alpha^{\omega})^{*}=0$, implying that the information coming from the neighbors with the same sensitive attribute as the anchor node is not used, which is expected to degrade the overall utility. 

{\color{mpurple}It always holds that $|R_{1}^{\chi}\alpha^{\chi} + R_{0}^{\chi}\alpha^{\chi} - 1| \geq 0$, where the equality is achieved when $\alpha^{\chi}= \frac{1}{R_{1}^{\chi} + R_{0}^{\chi}}$. Therefore, if $0 \leq \frac{1}{R_{1}^{\chi} + R_{0}^{\chi}} \leq \alpha_{max}^{\chi}$, the optimal solution becomes $(\alpha^{\chi})^{*}= \frac{1}{R_{1}^{\chi} + R_{0}^{\chi}}$. For the case, $\frac{1}{R_{1}^{\chi} + R_{0}^{\chi}} \geq \alpha_{max}^{\chi}$, the optimal solution is obtained on the boundary where $(\alpha^{\chi})^{*}= \alpha_{max}^{\chi}$}. Therefore, the optimal solution of the problem in \eqref{eq:optim} can be obtained in closed form as:
\begin{equation}
\label{eq:sol}
(\alpha^{\chi})^{*}=
    \begin{cases}
      \alpha^{\chi}_{max}, & \text{if}\ R_{1}^{\chi} + R_{0}^{\chi} < \frac{1}{\alpha^{\chi}_{max}}, \\
    \frac{1}{R_{1}^{\chi} + R_{0}^{\chi}}
      , & \text{else}.
    \end{cases}
\end{equation}

We design our fair attention layer such that the utilized attention coefficients satisfy the optimal amount of attention that should be assigned to the inter-edges, $(\alpha^{\chi})^{*}$, presented in \eqref{eq:sol}. Overall, the fair attention design and information aggregation process in FairGAT can be summarized as:
\begin{enumerate}
    \item  $e\left(\boldsymbol{h}_i^{l}, \boldsymbol{h}^{l}_j\right)=\operatorname{LReLU}\left((\boldsymbol{a}^{l})^{\top} \cdot\left[\boldsymbol{W}^{l} \boldsymbol{h}^{l}_i \| \boldsymbol{W}^{l} \boldsymbol{h}^{l}_j\right]\right)$,
   \vspace{0.1cm} 
\item \begin{equation}
\label{eq:fairatt}
   \hspace{-6mm} \alpha^{l}_{i j}\!=\!\!\begin{cases}
  (\alpha^{\chi})^{*} \frac{\exp \left(e\left(\boldsymbol{h}^{l}_i, \boldsymbol{h}^{l}_j\right)\right)}{\sum_{j^{\prime} \in \mathcal{N}_i \cap s_i \neq s_j} \exp \left(e\left(\boldsymbol{h}^{l}_i, \boldsymbol{h}^{l}_{j^{\prime}}\right)\right)},&\hspace{-1mm}  s_i \neq s_j  \\
   
    (\alpha^{\omega})^{*} \frac{\exp \left(e\left(\boldsymbol{h}^{l}_i, \boldsymbol{h}^{l}_j\right)\right)}{\sum_{j^{\prime} \in \mathcal{N}_i \cap s_i = s_j} \exp \left(e\left(\boldsymbol{h}^{l}_i, \boldsymbol{h}^{l}_{j^{\prime}}\right)\right)},              &\hspace{-1mm}  s_i = s_j
\end{cases}\end{equation}
  \vspace{0.1cm}
\item $\boldsymbol{h}_i^{l+1}=\sigma\left(\sum_{j \in \mathcal{N}_i} \alpha^{l}_{i j} \cdot \boldsymbol{W}^{l} \boldsymbol{h}^{l}_j\right)$.
\end{enumerate}
Note that step 2 in the proposed fair attention design ensures the optimal values of $(\alpha^{\chi})^{*}$ and $(\alpha^{\omega})^{*}$ in \eqref{eq:sol}. Furthermore, the proposed scheme differs from conventional attention learning in the employment of individual softmax activation functions over different sensitive groups, which does not add significant additional computational complexity over the conventional scheme (see Table \ref{table:runtime} for empirical run-time results). {Therefore, fair attention learning proposed herein provides an efficient solution for bias mitigation while also enjoying the flexible non-uniform weights assigned to different neighbors, similar to conventional GATs.}

\subsubsection{Spectral normalization}
As shown in \eqref{eq:main_res1} and \eqref{eq:main_res2}, the disparity between { the input representations, $\delta_{h}^{l}$, is multiplied by $\sigma_{max}(\mathbf{W}^{l})$ at every layer $l$. FairGAT ensures that this factor does not amplify the already existing bias by applying spectral normalization to $\mathbf{W}^{l}$, i.e., $\sigma_{max}( \operatorname{SN}(\mathbf{W}^{l}))=1$, where $\operatorname{SN}(\cdot)$ denotes a spectral normalization operator that applies to the input matrix.

The largest singular value, also known as the spectral norm, of a matrix $\mathbf{W} \in \mathbb{R}^{F_1 \times F_2}$ can be written as: 
\begin{equation}
    \sigma_{max}(\mathbf{W})=\max _{\boldsymbol{\xi} \in \mathbb{R}^{F_1}, \boldsymbol{\xi} \neq \mathbf{0}} \frac{\|\mathbf{W} \boldsymbol{\xi}\|_2}{\|\boldsymbol{\xi}\|_2}.
\end{equation}
Consider the input-output relation $\mathbf{\hat{y}}=\sigma\left(\boldsymbol{W} \mathbf{h}\right)$. Here, if a perturbation $\boldsymbol{\xi}$ is applied to the input, i.e., $\mathbf{\tilde{y}}=\sigma\left(\boldsymbol{W} (\mathbf{h} + \boldsymbol{\xi})\right)$, we have
\begin{equation}
\begin{split}
 \frac{\|\mathbf{\tilde{y}} - \mathbf{\hat{y}}\|_{2}}{\|\boldsymbol{\xi}\|_{2}} &= \frac{\| \sigma\left(\boldsymbol{W} \mathbf{h}\right) - \sigma\left(\boldsymbol{W} (\mathbf{h} + \boldsymbol{\xi})\right)\|_{2}}{\|\boldsymbol{\xi}\|_{2}}\\
 &\leq \frac{L\| \left(\boldsymbol{W} \mathbf{h}\right) - \left(\boldsymbol{W} (\mathbf{h} + \boldsymbol{\xi})\right)\|_{2}}{\|\boldsymbol{\xi}\|_{2}}\\
 &=\frac{L\| \boldsymbol{W} \boldsymbol{\xi})\|_{2}}{\|\boldsymbol{\xi}\|_{2}}\\
 &\leq L \sigma_{max}(\mathbf{W}).
  \end{split}
\end{equation}
Therefore, although limiting the spectral norm of weight matrices is primarily utilized to prevent bias amplification in our work, it can also help to improve the robustness and generalizability of neural networks \cite{yoshida2017spectral}.} In particular, spectral normalization can help FairGAT be more robust when the training and test data distributions do not match.

 \textbf{Remark. }It is important to note that, even though FairGAT applies spectral normalization to ensure $\sigma_{max}(\mathbf{W}^{l})=1$ at every layer $l$, $\sigma_{max}(\mathbf{W}^{l})$ can be further reduced for a lower upper bound on $\delta_{\hat{y}}$. For this purpose, a hyperparameter $\kappa \leq 1$ can be utilized to scale normalized weight matrices, as scaling a matrix would lead to the scaling of singular values (hence also the largest one), i.e., $\kappa \mathbf{W} = \sum_{i=1}^r \kappa \sigma_i \mathbf{u}_i \mathbf{v}_i^* $ (singular value decomposition of a weight matrix $\mathbf{W} \in \mathbb{R}^{F_1 \times F_2}$ can be written as $\mathbf{W} =\sum_{i=1}^r \sigma_i \mathbf{u}_i \mathbf{v}_i^*$, where $r \leq min(F_1, F_2)$). Note that the upper bound of $\delta_{\hat{y}}$ is minimized if $\sigma_{max}(\mathbf{W})=0$ by setting $\kappa=0$. However, such scaling would also prevent any learning, as the same predictions would be output by the model, providing perfect group fairness yet completely disregarding the utility. Thus, introducing $\kappa$ allows a trade-off between fairness and utility, and can potentially improve performance. However, we did not introduce such a hyperparameter to alleviate the parameter-tuning process. 

\subsubsection{Scaling representations}
Both Theorem \ref{theorem:gat} and Lemma \ref{lemma:mlp} demonstrate that the maximal deviation of the aggregated representations, $\Delta_{z}$, influences the disparity between the output representations of different sensitive groups. Furthermore, Theorem \ref{theorem:gat} suggests that the maximal deviation $\Delta_{c}$ is another factor in the disparity of attention layers. Motivated by these findings, FairGAT scales $\mathbf{Z}^{l+1}$ and $\mathbf{C}^{l+1}$ by a factor $\eta$ at every layer $l$, which also scales $\Delta_{z}$ and $\Delta_{c}$ by the same factor, i.e., $\left\|\eta \mathbf{z}^{l+1}_j-\eta \bar{\mathbf{z}}_{s}^{l+1}\right\|_{\infty} = \eta \left\|\mathbf{z}^{l+1}_j- \bar{\mathbf{z}}_{s}^{l+1}\right\|_{\infty}\leq (\eta \Delta_{z}^{(s)})^{l+1}$, $\forall v_{j} \in \mathcal{S}_{s}$. Here, $\eta$ is a hyperparameter utilized to provide a trade-off between the fairness and utility.


\begin{algorithm}
\caption{FairGAT}\label{alg:fairgat}
\KwData{$\mathcal{G}:=(\mathcal{V}, \mathcal{E})$, $\bbX$, $\bbs$, $\alpha_{max}^{\chi}$, $\eta$ }
\KwResult{$\hat{\mathbf{y}}$}
\begin{enumerate}[noitemsep]
    \item[S1.] Employ fair attention learning described in \eqref{eq:fairatt} at every attention layer
 \item[S2.] Apply spectral normalization to weight matrices $\mathbf{W}^{l}$ at every layer $l$ to ensure $\sigma_{max}( \operatorname{SN}(\mathbf{W}^{l}))=1$. 
\item[S3.]  Scale aggregated representations $\mathbf{Z}^{l+1}$ at every $l$ and $\mathbf{C}^{l+1}$ at every attention layer $l$ by a factor $\eta$
\end{enumerate}
\end{algorithm}

\section{Experiments}
\subsection{Datasets and Experimental Setup}



\noindent\textbf{Datasets.} The performance of the proposed FairGAT framework is evaluated on the node classification task over real-world social networks Pokec-z and Pokec-n \cite{say}, and the Recidivism graph \cite{bail}. Pokec-z and Pokec-n are social graphs that are sampled from a larger one, Pokec \cite{pokec}, which is a Facebook-like social network from Slovakia. Pokec-z and Pokec-n are sampled from an anonymized version of the Pokec network of 2012, where nodes correspond to users who live in two major regions, and the region information is utilized as the sensitive attribute \cite{say}. The working field of the users is binarized and utilized as the labels to be predicted in node classification. For building the Recidivism graph, the information of defendants (corresponding to nodes) who got released on bail at the U.S. state courts during 1990-2009 \cite{bail} is used, where the edges are created based on the affinity of past criminal records and demographics. Ethnicity of the defendants is used as the sensitive attribute for this graph, and the node classification task classifies defendants into bail (i.e., the defendant is not likely to commit a violent crime if released) or no bail (i.e., the defendant is likely to commit a violent crime if released) \cite{nifty}. 

Although we focused on node classification when building FairGAT in Section \ref{sec:fairgat}, we also consider link prediction as an alternative task for evaluation. For link prediction, experimental results are obtained over real-world citation networks Cora, Citeseer, and Pubmed. These citation networks consider articles as nodes and descriptions of articles as their nodal attributes. In these datasets, similar to the setups in \cite{dyadic,biased}, the category of the articles is used as the sensitive attribute for the evaluation of fairness-aware link prediction. Statistical information for all datasets is presented in Appendix \ref{app:data_stats}.

\noindent \textbf{Evaluation Metrics.} Classification accuracy is used to measure the utility for node classification. As fairness metrics, two quantitative measures of group fairness are used: \textit{statistical parity}, $\Delta_{S P}:=|P(\hat{c}_{j}=1 \mid s_{j}=0)-P(\hat{c}_{j}=1 \mid s_{j}=1)|$ \cite{sp}, and \textit{equal opportunity}, $\Delta_{E O}:=|P(\hat{c}_{j}=1 \mid y_{j}=1, s_{j}=0)-P(\hat{c}_{j}=1 \mid y_{j}=1, s_{j}=1)|$ \cite{eo}, where $y$ is the ground truth label, and $\hat{c}$ stands for the predicted binary class label. Statistical parity is a measure for the independence of positive rate from the sensitive attribute, and equal opportunity indicates the level of the independence of true positive rate from the sensitive attribute. For both metrics, lower values are desired.

For link prediction experiments, accuracy is again employed as the utility metric. For fairness evaluation, $\Delta DP_{m}$, $\Delta EO_{m}$, $\Delta DP_{g}$, $\Delta EO_{g}$, $\Delta DP_{s}$, and $\Delta EO_{s}$ that are introduced in \cite{spinelli2021fairdrop} are utilized. These metrics measure the demographic parity difference and equalized odd difference among multiple sensitive groups, where
$\Delta DP=\max_s E[\hat{Y}_{j} \mid e_{j} \in \mathcal{S}_{s}]-\min _s E[\hat{Y}_{j} \mid e_{j} \in \mathcal{S}_{s}]$ and $
\Delta EO=\max (\Delta \mathrm{TPR}, \Delta \mathrm{FPR})$ given
$$\begin{aligned}
\Delta \mathrm{TPR}:= & \max _s E[\hat{Y}_{j}=1 \mid e_{j} \in \mathcal{S}_{s}, Y_{j}=1]  -\min _s E[\hat{Y}_{j}=1 \mid e_{j} \in \mathcal{S}_{s}, Y_{j}=1] \\
\Delta \mathrm{FPR}:= & \max _s E[\hat{Y}_{j}=1 \mid e_{j} \in \mathcal{S}_{s}, Y_{j}=0]  -\min _s E[\hat{Y}_{j}=1 \mid e_{j} \in \mathcal{S}_{s}, Y_{j}=0].
\end{aligned}
$$
Here, $\hat{Y}_{j}$ is the prediction for the existence of the edge $e_{j}$ and $Y_{j}$ is the ground-truth label for whether $e_{j}$ exists in the input graph ($Y_{j}=1$) or not ($Y_{j}=0$). In the fairness measures, different subindices correspond to different sensitive group definitions. Specifically, subscripts m, g, and s represent the sensitive groups: mixed dyadic, group dyadic, and subgroup dyadic defined in \cite{spinelli2021fairdrop, masrour2020bursting}, respectively.

\begin{table}[h]
	\centering
\caption{Comparative results.}
\label{table:comp}
\begin{footnotesize}
\begin{tabular}{l c c c }
\toprule
                                                    
                  Pokec-z           & Accuracy ($\%$) & $\Delta_{S P}$ ($\%$) & $\Delta_{E O}$ ($\%$)  \\ \toprule
{GAT} 
                   & $  66.26 \pm 0.95$ & $3.63 \pm 2.56$  & $4.30 \pm 2.38$   \\ 
                   \midrule
{FairGNN} 
                   & $  \mathbf{67.71} \pm 0.70$ & $\mathbf{2.27} \pm 0.94$  & $2.31 \pm 1.01$    \\  
                  
{EDITS} 
                   & $  63.89 \pm 0.67$ & $3.27 \pm 1.99$  & $2.93 \pm 2.19$  \\ 
                 
{NIFTY} 
                   & $ 66.59 \pm 0.76$ & $4.21 \pm 1.38$  & $4.19 \pm 2.69$   \\ 
                   \midrule

{FairGAT}   & $  66.29 \pm 0.56$ & $2.55 \pm 0.50$  & $\mathbf{1.63} \pm 0.88$  \\

                    \toprule  \\                             
                  Pokec-n           & Accuracy ($\%$) & $\Delta_{S P}$ ($\%$) & $\Delta_{E O}$ ($\%$)  \\ \toprule
{GAT} 
                   &  $  67.50 \pm 0.43$ & $2.26 \pm 2.82$  & $3.56 \pm 1.73$  \\ 
                   \midrule
{FairGNN} 
                   &  $  65.81 \pm 0.84$ & $2.21 \pm 1.38$  & $2.97 \pm 1.27$   \\  
                  
{EDITS} 
                   &  $  63.47 \pm 0.92$ & $2.01 \pm 1.54$  & $2.48 \pm 2.33$    \\ 
                  
{NIFTY} 
                      & $  \mathbf{68.41} \pm 1.52$ & $1.41 \pm 0.72$  & $2.30 \pm 1.85$   \\ 
                   \midrule

{FairGAT}    &    $67.81 \pm 1.09$ &  $\mathbf{0.71} \pm 0.74 $ &   $\mathbf{1.23} \pm 0.58$  \\
\toprule
                                                   
               \\   Recidivism           & Accuracy ($\%$) & $\Delta_{S P}$ ($\%$) & $\Delta_{E O}$ ($\%$)  \\ \toprule
{GAT} 
                   & $  \mathbf{95.63} \pm 0.20$ & $8.08 \pm 1.69$  & $2.09 \pm 1.05$  \\ 
                   \midrule
{FairGNN} 
                   & $  95.18 \pm 0.25 $ & $7.31 \pm 1.88$  & $1.27 \pm 0.98$  \\  
                 
{EDITS} 
                   & $  88.52 \pm 0.63$ & $\mathbf{6.59} \pm 2.07$  & $1.73 \pm 1.07$    \\ 
                  
{NIFTY} 
                  & $  88.52 \pm 2.35 $ & $6.74 \pm 2.36$  & $ 1.48 \pm 1.92$    \\ 
                   \midrule

{FairGAT}   &    $94.93 \pm 0.15$ &  $7.39 \pm 1.76 $ &   $\mathbf{1.02} \pm 0.05$ \\
                                                          \bottomrule
                                                          
\end{tabular}
\end{footnotesize}
\end{table}
\begin{table}[h]
	\centering
\caption{Ablation study.}
\label{table:ablation}
\begin{footnotesize}
\begin{tabular}{l c c c }
\toprule
                                                    
                  Pokec-z           & Accuracy ($\%$) & $\Delta_{S P}$ ($\%$) & $\Delta_{E O}$ ($\%$)  \\ \toprule
{GAT} 
                   & $  66.26 \pm 0.95$ & $3.63 \pm 2.56$  & $4.30 \pm 2.38$   \\ 
                   \midrule
{Steps 1\&2} 
                   & $  66.84 \pm 0.28$ & $3.93 \pm 2.81$  & $4.85 \pm 2.71$    \\ 
{Steps 1\&3} 
                   & $  66.74 \pm 0.84$ & $\mathbf{2.10} \pm 1.54$  & $2.21 \pm 1.78$       \\ 
                       
{Steps 2\&3} 
                   & $  \mathbf{67.78} \pm 0.58$ & $5.59 \pm 1.12$  & $5.69 \pm 1.48$    \\ 
                   \midrule
{FairGAT}   & $  66.29 \pm 0.56$ & $2.55 \pm 0.50$  & $\mathbf{1.63} \pm 0.88$  \\

                    \toprule  \\                             
                  Pokec-n           & Accuracy ($\%$) & $\Delta_{S P}$ ($\%$) & $\Delta_{E O}$ ($\%$)  \\ \toprule
{GAT} 
                   &  $  67.50 \pm 0.43$ & $2.26 \pm 2.82$  & $3.56 \pm 1.73$  \\ 
                   \midrule

{Steps 1\&2} 
                     & $  64.60 \pm 0.74$ & $4.37 \pm 1.71$  & $4.29 \pm 3.35$    \\
{Steps 1\&3} 
                  & $  66.91 \pm 0.98$ & $1.95 \pm 1.46$  & $3.64 \pm 2.64$  \\ 
                 
{Steps 2\&3} 
                      & $  \mathbf{67.59} \pm 0.84$ & $1.57 \pm 1.12$  & $2.27 \pm 1.03$   \\ 
                   \midrule
{FairGAT}    &    $67.14 \pm 1.06$ &  $\mathbf{0.90} \pm 0.61 $ &   $\mathbf{1.68} \pm 1.30$  \\
\toprule
                                                   
               \\   Recidivism           & Accuracy ($\%$) & $\Delta_{S P}$ ($\%$) & $\Delta_{E O}$ ($\%$)  \\ \toprule
{GAT} 
                   & $  \mathbf{95.63} \pm 0.20$ & $8.08 \pm 1.69$  & $2.09 \pm 1.05$  \\ 
                   \midrule
{Steps 1\&2} 
                  & $  89.95 \pm 0.41$ & $\mathbf{7.06} \pm 2.22$  & $1.81 \pm 2.09$   \\ 
                  
{Steps 1\&3} 
                   & $  95.23 \pm 0.17$ & $8.12 \pm 2.13$  & $1.44 \pm 0.52$   \\ 
                 
{Steps 2\&3} 
                   & $  95.38 \pm 0.20$ & $7.94 \pm 1.67$  & $1.71 \pm 1.02$    \\ 
                   \midrule
{FairGAT}   &    $94.93 \pm 0.15$ &  $7.39 \pm 1.76 $ &   $\mathbf{1.02} \pm 0.05$ \\
                                                          \bottomrule
                                                          
\end{tabular}
\end{footnotesize}
\end{table}
\noindent \textbf{Implementation details.}  
\label{subsec:implementation}
In the experiments, a network that consists of two attention layers (conventional GAT layers for the baselines) and one fully connected layer is trained in a supervised manner for node classification, where each attention layer is followed by a ReLU activation. This structure is kept the same for all baselines as well as FairGAT for a fair performance comparison. The model is trained over $40\%$ of the nodes, while the remaining nodes are equally divided to validation and test sets. The test-set performance of the model that performs the best on the validation set is reported. For all experiments, results are collected for five random data splits, and their averages and standard deviations are reported. 

For link prediction, the experimental setting in \cite{spinelli2021fairdrop} is kept the same, where a two-layer attention network (consists of conventional GAT layers for the baseline) is trained for supervised link prediction. The hyperparameters of FairGAT and all other baselines are tuned via a grid search on cross-validation sets, see Appendix \ref{app:hypers} for the utilized hyperparameter values in all experiments.

\noindent \textbf{Baselines.}
Herein, we present the performances of three fairness-aware baseline studies for node classification: FairGNN \cite{say}, EDITS \cite{dong2022edits}, and NIFTY \cite{nifty}. For improving fairness in a supervised setting, FairGNN \cite{say} employs adversarial debiasing and a covariance-based regularizer (the absolute covariance between the sensitive attribute and estimated labels $\hat{\mathbf{c}}$) together. It is important to note that FairGNN is applicable to only supervised learning settings due to its covariance-based regularizer's design, while FairGAT can be utilized in both unsupervised and supervised learning frameworks. EDITS \cite{dong2022edits} is a model-agnostic debiasing framework that mitigates the bias in attributed networks before they are fed into any GNN. Specifically, it creates debiased versions of the nodal attributes and the graph structure, which are then input to the GAT network for node classification for the results in this work. Finally, NIFTY \cite{nifty} employs a layer-wise weight normalization scheme along with a fair graph augmentation. Its objective function consists of both supervised and unsupervised components, where the classification loss of the GAT network is employed to be the supervised loss. For link prediction, FairDrop \cite{spinelli2021fairdrop} is utilized as the fairness-aware baseline, which applies an edge manipulation on the input graph to mitigate structural bias.

\subsection{Results for Node Classification}

The results of node classification are presented in Table \ref{table:comp} in terms of fairness and utility metrics for both FairGAT and baselines. In Table \ref{table:comp}, ``GAT" represents the natural baseline where the conventional GAT layers \cite{gat} are employed for node classification without any spectral normalization (step 2 in FairGAT algorithm) or representation scaling (step 3 in FairGAT algorithm). Moreover, ``FairGNN" \cite{say}, ``EDITS" \cite{dong2022edits} and ``NIFTY" \cite{nifty} are all fairness-aware baseline studies. 

The results in Table \ref{table:comp} demonstrate that FairGAT significantly improves the naive baseline, GAT, in terms of fairness metrics, while yielding similar utility. Specifically, FairGAT achieves $30\%$ to $60\%$ improvement in all fairness measures for every dataset compared to GAT, except for $\Delta_{SP}$ for Recidivism. Furthermore, FairGAT consistently outperforms every fairness-aware baseline in terms of $\Delta_{EO}$ on all datasets. While FairGAT also leads to the best $\Delta_{SP}$ value on Pokec-n, FairGNN results in a better performance in terms of $\Delta_{SP}$ on Pokec-z. However, the fairness improvements provided by FairGNN are observed to vary over different datasets (e.g., it is the worst-performing fairness-aware baseline on Pokec-n), which can be explained by the instability issues related to adversarial training \cite{kodali2017convergence}. On the Recidivism graph, all other fairness-aware baselines result in better or similar $\Delta_{SP}$ values compared to FairGAT. However, for NIFTY and EDITS, the superior $\Delta_{SP}$ performance on the Recidivism dataset is accompanied by a considerable decrease in classification accuracy. Furthermore, it can be observed that FairGAT leads to the lowest standard deviation values for fairness measures for all datasets, hence provides a better robustness in terms of fairness. Overall, the results demonstrate that FairGAT generally improves the fairness measures and consistently achieves better stability in terms of fairness compared to other state-of-the-art fairness-aware baselines, while providing similar utility to the conventional GAT network.

 An ablation study is also provided in Table \ref{table:ablation} in order to demonstrate the influences of different steps in Algorithm \ref{alg:fairgat}. In Table \ref{table:ablation}, ``Step 1" stands for the employment of fair attention layers, as described in \eqref{eq:fairatt}. ``Step 2" and ``Step 3" represent spectral normalization of weight matrices and hidden representation scaling that are detailed in Subsection \ref{subsec:meth}, respectively. Overall, the ablation study signifies that FairGAT typically achieves the best fairness measures, together with similar or better utility, compared to a framework which lacks one of the steps in Algorithm \ref{alg:fairgat}. Therefore, the ablation study suggests that all steps in FairGAT are essential for the success and robustness of the algorithm. 
\setcounter{table}{3}
\begin{table}[h]
	\centering
\caption{Run-time comparison}
\label{table:runtime}
\begin{footnotesize}

\begin{tabular}{l c c c }
\toprule
   Epoch (sec)                                               & {Pokec-z}& {Pokec-n} & Recidivism                                     \\ 
 \midrule
{GAT} & {$0.30$} & $0.23$  & $3.07$ \\ \midrule
{FairGNN} & {$0.68$} & $0.50$  & $6.91$ \\
{EDITS} & {$0.74$} & $0.41$  & $9.31$ \\
{NIFTY} & {$1.28$} & $0.91$  & $13.97$ \\ \midrule
{FairGAT} & {$0.31$} & $0.24$  & $3.11$
\\ \bottomrule
\end{tabular}
\end{footnotesize}
\end{table}

In order to demonstrate the time complexity incurred by the proposed framework, Table \ref{table:runtime} presents the average run-time of each epoch for FairGAT and the baselines. Note that the run-time for EDITS \cite{dong2022edits} is obtained by excluding the time incurred by the fairness-aware pre-processing step. For EDITS, the table shows the average run-time of each epoch for the training of GAT network with pre-processed inputs. The results confirm our claim that the proposed fair attention design does not add significant additional computational complexity over the conventional GAT layers. Furthermore, the results also demonstrate that FairGAT can provide a more efficient solution to combat bias compared to other fairness-aware baselines. 
\begin{table*}[t]
	\centering
\caption{Comparative results of FairGAT for link prediction.}
\label{table:fairlp}
\begin{scriptsize}
\begin{tabular}{l c c c c c c c c c}
\toprule
                        &     & Accuracy ($\%$) &  $\Delta DP_{m}$ ($\%$) & $\Delta EO_{m}$ ($\%$) &  $\Delta DP_{g}$ ($\%$) & $\Delta EO_{g}$ ($\%$) &  $\Delta DP_{s}$ ($\%$) & $\Delta EO_{s}$ ($\%$) \\ \midrule
\multirow{2}{1.418cm}[1.5ex]{\textbf{Cora}} 
 & {FairDrop}   & $74.94 \pm 1.21$ &   $\mathbf{40.84} \pm 2.73$ &  $25.48 \pm  2.04$ &  $17.99 \pm  2.61$ &  $25.56 \pm  3.69$ &  $85.06 \pm  3.56$ &  $100.00 \pm  0.00$ \\ 
&{FairGAT} & $74.73 \pm 1.34$ & $\mathbf{40.37} \pm 2.80$ &  $\mathbf{17.83} \pm  2.54$ &  $\mathbf{12.97} \pm  4.82$ &  $\mathbf{17.55} \pm  3.62$ &  $\mathbf{73.19} \pm  7.64$ &  $\mathbf{96.26} \pm  7.33$ \\ \midrule
\cmidrule(r){2-5} 
\multirow{2}{1.418cm}[1.5ex]{\textbf{Citeseer}}
&{FairDrop}   & $65.98 \pm 1.57$ & $23.80 \pm 2.87$ &  $12.50 \pm  4.26$ &  $32.05 \pm  5.74$ &  $47.32 \pm  7.72$ &  $65.37 \pm  3.87$ &  $73.81 \pm  5.82$  \\ 
&{FairGAT} & $\mathbf{69.06} \pm 0.91$ & $25.50 \pm 2.24$ &  $\mathbf{8.67} \pm  2.28$ &  $\mathbf{24.28} \pm  5.56$ &  $\mathbf{31.26} \pm  5.03$ &  $55.14 \pm  6.48$ &  $\mathbf{60.58} \pm  6.60$ \\ \midrule
\cmidrule(r){2-5} 
\multirow{2}{1.418cm}[1.5ex]{\textbf{Pubmed}} 
&{FairDrop}   & $75.09 \pm 0.45$ & $34.04 \pm 0.97$ &  $17.99 \pm  0.95$ &  $\mathbf{5.96} \pm  1.08$ &  $\mathbf{6.38} \pm  1.84$ &  $41.89 \pm  1.42$ &  $36.08 \pm  2.29$  \\ 
&{FairGAT} & $75.39 \pm 0.87$ & $\mathbf{28.68} \pm 1.31$ &  $\mathbf{7.92} \pm  0.80$ &  $7.10 \pm  0.66$ &  $9.38 \pm  0.75$ &  $\mathbf{35.20} \pm  2.13$ &  $\textbf{17.87} \pm  2.24$ \\ \midrule

\end{tabular}
\end{scriptsize}
\end{table*}

\subsection{Results for Link Prediction}
Although our analysis is developed for node classification and for a binary-sensitive attribute, we also obtain experimental results for link prediction. For this task, the utilized datasets Cora, Citeseer, and Pubmed contain non-binary sensitive attributes, for which we still employ the fair attention described in \eqref{eq:fairatt} by directly tuning $(\alpha^{\chi})^{*}$ as a hyperparamater. The results for link prediction are presented in Table \ref{table:fairlp}, where ``FairDrop" stands for \cite{spinelli2021fairdrop}, which is also fairness-aware. The results in Table \ref{table:fairlp} demonstrate that FairGAT typically outperforms FairDrop in terms of fairness measures while providing similar or better utility. Overall, the experimental results signify that FairGAT shows promising results also for the link prediction task and non-binary sensitive attributes.

\section{Conclusion and Future Work}
This study presents a fairness-aware graph-based learning framework, FairGAT, which leverages a novel attention learning strategy that can mitigate bias. The design of the proposed scheme is based on a theoretical analysis illuminating the sources of bias in a GAT-based neural network trained for node classification. The fair attention design in FairGAT incurs negligible additional computational complexity compared to the conventional GAT layer, and it can be flexibly employed with other fairness enhancement strategies. Experiments on real-world networks for node classification demonstrate that FairGAT typically provides better fairness measures together with similar utility compared to the state-of-the-art fairness-aware baselines. Furthermore, the obtained results for link prediction show the promising fairness performance of FairGAT also for link prediction and non-binary sensitive attributes.

The present work opens up a number of possible
future directions: (i) extension of the present analysis to the case where multiple, non-binary sensitive attributes are available; (ii) the consideration of different aggregation schemes in bias analysis in addition to mean-aggregation.

\bibliographystyle{IEEEtran}
\bibliography{refs}


\appendix
\onecolumn
\section{Lemma \ref{lemma:activated} and Its Proof}
\label{app:lemma_activated}
\begin{lemma}
\label{lemma:activated}
The disparity between the aggregated representations $\mathbf{Z} \in \mathbb{R}^{N \times F}$ from different sensitive groups is related to the disparity between the hidden representations $\mathbf{H}:= \sigma(\mathbf{Z}) \in \mathbb{R}^{N \times F}$ from different sensitive groups as follows:
\begin{equation}
\begin{split}
\delta_{h}:=\left\|\operatorname{mean}(\mathbf{h}_{j} \mid s_{j}=0) - \operatorname{mean}(\mathbf{h}_{j} \mid s_{j}=1)\right\|_{2} &= \left\|\frac{1}{|\mathcal{S}_{0}|} \sum_{v_j \in \mathcal{S}_{0}} \mathbf{h}_{j}-\frac{1}{|\mathcal{S}_{1}|} \sum_{v_j \in \mathcal{S}_{1}} \mathbf{h}_{j}\right\|_{2}\\
&\leq  L\left( \left\|\frac{1}{|\mathcal{S}_{0}|} \sum_{v_j \in \mathcal{S}_{0}} \mathbf{z}_{j}-\frac{1}{|\mathcal{S}_{1}|} \sum_{v_j \in \mathcal{S}_{1}} \mathbf{z}_{j}\right\|_{2} + 2 \sqrt{N} \Delta_{z} \right).
\end{split}
\end{equation}
Here, $L$ is the Lipschitz constant of nonlinear activation $\sigma$, and maximal deviation $\Delta_{z}$ is defined to be $\Delta_{z}:= \operatorname{max}(\Delta^{(0)}_{z}, \Delta^{(1)}_{z})$, where $\left\|\mathbf{z}_j-\bar{\mathbf{z}}_{s}\right\|_{\infty} \leq \Delta^{(s)}_{z}$, $\forall v_j \in \mathcal{S}_{s}$ with $\bar{\mathbf{z}}_{s} = \frac{1}{|\mathcal{S}^{s}|} \sum_{v_j \in \mathcal{S}_{s}} \mathbf{z}_{j}$ for $s=0,1$.

\end{lemma}

\textbf{Proof of Lemma \ref{lemma:activated}: } Note that as this analysis applies to every layer in the same way, we drop the superscript $l$ used to denote the layer. The disparity between the representations from different sensitive groups follow as:
    
    \begin{equation}
    \label{eq:defs}
        \|\frac{1}{|\mathcal{S}_{0}|} \sum_{v_j \in \mathcal{S}_{0}} \mathbf{h}_{j}-\frac{1}{|\mathcal{S}_{1}|} \sum_{v_j \in \mathcal{S}_{1}} \mathbf{h}_{j}\|_{2}= \|\frac{1}{|\mathcal{S}_{0}|} \sum_{v_j \in \mathcal{S}_{0}} \sigma(\mathbf{z}_{j})-\frac{1}{|\mathcal{S}_{1}|} \sum_{v_j \in \mathcal{S}_{1}}\sigma(\mathbf{z}_{j})\|_{2}
    \end{equation}
    
We can write $\mathbf{z}_{j}= \bar{\mathbf{z}}_{s} + \boldsymbol{\delta}^{(s)}_{j}$, $\forall v_j \in \mathcal{S}_{s}$ , where $\bar{\mathbf{z}}_{s} = \frac{1}{|\mathcal{S}_{s}|} \sum_{v_j \in \mathcal{S}_{s}} \mathbf{z}_{j}$ for $s=0,1$. If the activation function $\sigma(.)$ is Lipschitz continuous with Lipschitz constant $L$ (applies to several nonlinear activations, such as rectified linear unit (ReLU), sigmoid), the following holds:
\begin{equation}
\label{eq:main_ineq}
\begin{split}
\operatorname{\sigma}((\bar{\mathbf{z}}_{0})_{i}) - L|\delta^{(0)}_{i,j}|  \leq \operatorname{\sigma}(z_{i,j})&= \operatorname{\sigma}((\bar{\mathbf{z}}_{0})_{i} + \delta^{(0)}_{i,j})\\
&\leq \operatorname{\sigma}((\bar{\mathbf{z}}_{0})_{i}) + L |\delta^{(0)}_{i,j}|, \forall i=1, \cdots F, \forall v_j \in \mathcal{S}_{0} \\
  \operatorname{\sigma}(\bar{\mathbf{z}}_{0}) - L|\boldsymbol{\delta}^{(0)}_{j}|  \preccurlyeq \operatorname{\sigma}(\mathbf{z}_{j})&= \operatorname{\sigma}(\bar{\mathbf{z}}_{0} + \boldsymbol{\delta}^{(0)}_{j})\\
  &\preccurlyeq \operatorname{\sigma}(\bar{\mathbf{z}}_{0}) + L|\boldsymbol{\delta}^{(0)}_{j}|, \forall v_j \in \mathcal{S}_{0}
  \end{split}
\end{equation}
where $|.|$ takes the element-wise absolute value of the input. The same inequalities can also be written for $\mathcal{S}_{1}$:
\begin{equation}
\label{eq:main_ineq2}
\begin{split}
 \operatorname{\sigma}(\bar{\mathbf{z}}_{1}) - L|\boldsymbol{\delta}^{(1)}_{j}|  \preccurlyeq \operatorname{\sigma}(\mathbf{z}_{j})= \operatorname{\sigma}(\bar{\mathbf{z}}_{1} + \boldsymbol{\delta}^{(1)}_{j})
  \preccurlyeq \operatorname{\sigma}(\bar{\mathbf{z}}_{1}) + L|\boldsymbol{\delta}^{(1)}_{j}|,\\ \forall v_j \in \mathcal{S}_{1}.
\end{split}
\end{equation}
Based on Equations \eqref{eq:defs}, \eqref{eq:main_ineq}, and \eqref{eq:main_ineq2}, following holds:
\begin{equation}
\begin{split}
\frac{1}{|\mathcal{S}_{0}|} \sum_{v_j \in \mathcal{S}_{0}} &\left(\operatorname{\sigma}(\bar{\mathbf{z}}_{0}) - L|\boldsymbol{\delta}^{(0)}_{j}| \right ) - \frac{1}{|\mathcal{S}_{1}|} \sum_{v_j \in \mathcal{S}_{1}}  \left(\operatorname{\sigma}(\bar{\mathbf{z}}_{1}) + L|\boldsymbol{\delta}^{(1)}_{j}| \right) \preccurlyeq \frac{1}{|\mathcal{S}_{0}|} \sum_{v_j \in \mathcal{S}_{0}} \mathbf{h}_{j}-\frac{1}{|\mathcal{S}_{1}|} \sum_{v_j \in \mathcal{S}_{1}} \mathbf{h}_{j}\\
&\preccurlyeq \frac{1}{|\mathcal{S}_{0}|} \sum_{v_j \in \mathcal{S}_{0}} \left( \operatorname{\sigma}(\bar{\mathbf{z}}_{0}) + L|\boldsymbol{\delta}^{(0)}_{j}|\right) - \frac{1}{|\mathcal{S}_{1}|} \sum_{v_j\in \mathcal{S}_{1}}  \left(\operatorname{\sigma}(\bar{\mathbf{z}}_{1}) - L|\boldsymbol{\delta}^{(1)}_{j}|  \right)
\end{split}
\end{equation}

\begin{equation}
\label{eq:before_norm}
\begin{split}
\operatorname{\sigma}(\bar{\mathbf{z}}_{0}) &- \operatorname{\sigma}(\bar{\mathbf{z}}_{1}) - \frac{1}{|\mathcal{S}_{0}|} \sum_{v_j \in \mathcal{S}_{0}} L|\boldsymbol{\delta}^{(0)}_{j}| - \frac{1}{|\mathcal{S}_{1}|} \sum_{v_j \in \mathcal{S}_{1}}  L|\boldsymbol{\delta}^{(1)}_{j}|  \preccurlyeq  \frac{1}{|\mathcal{S}_{0}|} \sum_{v_j \in \mathcal{S}_{0}} \mathbf{h}_{j}-\frac{1}{|\mathcal{S}_{1}|} \sum_{v_j \in \mathcal{S}_{1}} \mathbf{h}_{j}\\
&\preccurlyeq \operatorname{\sigma}(\bar{\mathbf{z}}_{0}) - \operatorname{\sigma}(\bar{\mathbf{z}}_{1}) + \frac{1}{|\mathcal{S}_{0}|} \sum_{v_j \in \mathcal{S}_{0}} L|\boldsymbol{\delta}^{(0)}_{j}| + \frac{1}{|\mathcal{S}_{1}|} \sum_{v_j \in \mathcal{S}_{1}}  L|\boldsymbol{\delta}^{(1)}_{j}|
\end{split}
\end{equation}
Define $\mathbf{a}:=\operatorname{\sigma}(\bar{\mathbf{z}}_{0}) - \operatorname{\sigma}(\bar{\mathbf{z}}_{1}) - \frac{1}{|\mathcal{S}_{0}|} \sum_{v_j \in \mathcal{S}_{0}} L|\boldsymbol{\delta}^{(0)}_{j}| - \frac{1}{|\mathcal{S}_{1}|} \sum_{v_j \in \mathcal{S}_{1}}  L|\boldsymbol{\delta}^{(1)}_{j}| $ and $\mathbf{b}:= \operatorname{\sigma}(\bar{\mathbf{z}}_{0}) - \operatorname{\sigma}(\bar{\mathbf{z}}_{1}) + \frac{1}{|\mathcal{S}_{0}|} \sum_{v_j \in \mathcal{S}_{0}} L|\boldsymbol{\delta}^{(0)}_{j}| + \frac{1}{|\mathcal{S}^{1}|} \sum_{v_j \in \mathcal{S}_{1}}  L|\boldsymbol{\delta}^{(1)}_{j}|$. Let $\bar{\mathbf{h}}_{s}$ denote $ \frac{1}{|\mathcal{S}_{s}|} \sum_{v_j \in \mathcal{S}_{s}} \mathbf{h}_{j}$ for $s=0,1$. Then, equation \eqref{eq:before_norm} leads to:
\begin{equation}
    |(\bar{\mathbf{h}}_{0})_{i}-(\bar{\mathbf{h}}_{1})_{i}| \leq \operatorname{max}(|a_{i}|,|b_{i}|), \forall i=1,\cdots, F.
    \end{equation}
If we consider the case, $|a_{i}| \geq |b_{i}| $. Then:
\begin{equation}
\begin{split}
    |(\bar{\mathbf{h}}_{0})_{i}-(\bar{\mathbf{h}}_{1})_{i}| &\leq \left|\operatorname{\sigma}((\bar{\mathbf{z}}_{0})_{i}) - \operatorname{\sigma}((\bar{\mathbf{z}}_{1})_{i}) - \frac{1}{|\mathcal{S}_{0}|} \sum_{v_j \in \mathcal{S}_{0}} L|\delta^{(0)}_{j,i}| - \frac{1}{|\mathcal{S}_{1}|} \sum_{v_j \in \mathcal{S}_{1}}  L|\delta^{(1)}_{j,i}|\right|\\
    &\leq \left|\operatorname{\sigma}((\bar{\mathbf{z}}_{0})_{i}) - \operatorname{\sigma}((\bar{\mathbf{z}}_{1})_{i})\right| + \left|\frac{1}{|\mathcal{S}_{0}|} \sum_{v_j \in \mathcal{S}_{0}} L|\delta^{(0)}_{j,i}| \right| + \left|\frac{1}{|\mathcal{S}_{1}|} \sum_{v_j \in \mathcal{S}_{1}}  L|\delta^{(1)}_{j,i}| \right|\\
    &\leq \left|\operatorname{\sigma}((\bar{\mathbf{z}}_{0})_{i}) - \operatorname{\sigma}((\bar{\mathbf{z}}_{1})_{i}) \right| + L\left| \Delta^{(0)} \right| +L \left| \Delta^{(1)} \right|,
    \end{split}
\end{equation}
where $\Delta^{(0)}_{i}:= \max_{j}|\delta^{(0)}_{j,i}|$, $\Delta^{(1)}_{i}:= \max_{j}|\delta^{(1)}_{j,i}|$ and $\Delta^{(0)}:=\|\boldsymbol{\Delta}^{(0)}\|_{\infty}$ and $\Delta^{(1)}:=\|\boldsymbol{\Delta}^{(1)}\|_{\infty}$.

Consider the term $\operatorname{\sigma}((\bar{\mathbf{z}}_{0})_{i}) - \operatorname{\sigma}((\bar{\mathbf{z}}_{1})_{i})$:
\begin{equation}
  \operatorname{\sigma}((\bar{\mathbf{z}}_{0})_{i}) - \operatorname{\sigma}((\bar{\mathbf{z}}_{1})_{i}) =  \operatorname{\sigma}((\bar{\mathbf{z}}_{0})_{i} + (\bar{\mathbf{z}}_{1})_{i} - (\bar{\mathbf{z}}_{1})_{i})  - \operatorname{\sigma}((\bar{\mathbf{z}}_{1})_{i}). 
\end{equation}
Utilizing Equations \eqref{eq:main_ineq} and \eqref{eq:main_ineq2}, $\operatorname{\sigma}((\bar{\mathbf{z}}_{0})_{i} + (\bar{\mathbf{z}}_{1})_{i} - (\bar{\mathbf{z}}_{1})_{i})  - \operatorname{\sigma}((\bar{\mathbf{z}}_{1})_{i})$ can be bounded by below and above:
\begin{equation}
\begin{split}
\operatorname{\sigma}((\bar{\mathbf{z}}_{1})_{i}) - L|(\bar{\mathbf{z}}_{0})_{i} - (\bar{\mathbf{z}}_{1})_{i}|  - \operatorname{\sigma}((\bar{\mathbf{z}}_{1})_{i})
&\leq \operatorname{\sigma}((\bar{\mathbf{z}}_{0})_{i} + (\bar{\mathbf{z}}_{1})_{i} - (\bar{\mathbf{z}}_{1})_{i})  - \operatorname{\sigma}((\bar{\mathbf{z}}_{1})_{i}) \\
&\leq \operatorname{\sigma}((\bar{\mathbf{z}}_{1})_{i}) + L|(\bar{\mathbf{z}}_{0})_{i} - (\bar{\mathbf{z}}_{1})_{i}|  - \operatorname{\sigma}((\bar{\mathbf{z}}_{1})_{i})
\end{split}
\end{equation}
   \begin{equation}
   \begin{split}
- L|(\bar{\mathbf{z}}_{0})_{i} - (\bar{\mathbf{z}}_{1})_{i}| \leq \operatorname{\sigma}((\bar{\mathbf{z}}_{0})_{i} + (\bar{\mathbf{z}}_{1})_{i} &- (\bar{\mathbf{z}}_{1})_{i})  - \operatorname{\sigma}((\bar{\mathbf{z}}_{1})_{i}) \leq L|(\bar{\mathbf{z}}_{0})_{i} - (\bar{\mathbf{z}}_{1})_{i}|
   \end{split}
\end{equation} 
   \begin{equation}
\left| \operatorname{\sigma}((\bar{\mathbf{z}}_{0})_{i}) - \operatorname{\sigma}((\bar{\mathbf{z}}_{1})_{i}) \right|  \leq
L|(\bar{\mathbf{z}}_{0})_{i} - (\bar{\mathbf{z}}_{1})_{i}|.
\end{equation} 
Therefore:
\begin{equation}
\label{eq:first_case}
    |(\bar{\mathbf{h}}_{0})_{i}-(\bar{\mathbf{h}}_{1})_{i}| \leq L|(\bar{\mathbf{z}}_{0})_{i} - (\bar{\mathbf{z}}_{1})_{i}| + L\left| \Delta^{(0)} \right| + L\left| \Delta^{(1)} \right|, \forall i \text{ such that } |a_{i}| \geq |b_{i}|. 
\end{equation}

Next step is to consider the case, $|a_{i}| < |b_{i}| $. For this case, following inequalities hold:
\begin{equation}
\label{eq:second_case}
\begin{split}
    |(\bar{\mathbf{h}}_{0})_{i}-(\bar{\mathbf{h}}_{1})_{i}| &\leq \left|\operatorname{\sigma}((\bar{\mathbf{z}}_{0})_{i}) - \operatorname{\sigma}((\bar{\mathbf{z}}_{1})_{i}) + \frac{1}{|\mathcal{S}_{0}|} \sum_{v_j \in \mathcal{S}_{0}} L|\delta^{(0)}_{j,i}| + \frac{1}{|\mathcal{S}_{1}|} \sum_{v_j \in \mathcal{S}_{1}}  L|\delta^{(1)}_{j,i}|\right|\\
    &\leq \left|\operatorname{\sigma}((\bar{\mathbf{z}}_{0})_{i}) - \operatorname{\sigma}((\bar{\mathbf{z}}_{1})_{i})\right| + \left|\frac{1}{|\mathcal{S}_{0}|} \sum_{v_j \in \mathcal{S}_{0}} L|\delta^{(0)}_{j,i}| \right| + \left|\frac{1}{|\mathcal{S}_{1}|} \sum_{v_j \in \mathcal{S}_{1}}  L|\delta^{(1)}_{j,i}| \right|\\
    &\leq \left|\operatorname{\sigma}((\bar{\mathbf{z}}_{0})_{i}) - \operatorname{\sigma}((\bar{\mathbf{z}}_{1})_{i}) \right| + L\left| \Delta^{(0)} \right| +L \left| \Delta^{(1)} \right|\\
    & \leq  L|(\bar{\mathbf{z}}_{0})_{i} - (\bar{\mathbf{z}}_{1})_{i}| + L\left| \Delta^{(0)} \right| + L\left| \Delta^{(1)} \right|, \forall i \text{ such that } |a_{i}| \geq |b_{i}|. 
\end{split}
\end{equation}
Combining Equations \eqref{eq:first_case} and \eqref{eq:second_case} and defining $\Delta_{z}:=\operatorname{max}(\Delta^{(0)}, \Delta^{(1)})$, the following inequality can be written:
\begin{equation}
   |(\bar{\mathbf{h}}_{0})_{i}-(\bar{\mathbf{h}}_{1})_{i}| \leq  L|(\bar{\mathbf{z}}_{0})_{i} - (\bar{\mathbf{z}}_{1})_{i}| + 2L\left| \Delta_{z} \right| , \forall i=1, \dots, F.
\end{equation}
which concludes:
\begin{equation}
   \|\bar{\mathbf{h}}_{0}-\bar{\mathbf{h}}_{1}\|_2  \leq L\left( \|\mathbf{\bar{z}}_{0} - \mathbf{\bar{z}}_{1}\|_{2} + 2 \sqrt{N} \Delta_{z} \right).
\end{equation}

\section{Proof of Theorem \ref{theorem:gat}}
\label{app:theorem_proof}

Here, without loss of generality, we will consider the $l$th GAT layer, where the input representations are denoted by $\mathbf{H}^{l}$ and output representations are $\mathbf{H}^{l+1}$. The disparity between the output representations follows as:
\begin{equation}
   \delta_{h}^{l+1}:=\left\|\operatorname{mean}(\mathbf{h}^{l+1}_{j} \mid s_{j}=0) - \operatorname{mean}(\mathbf{h}^{l+1}_{j} \mid s_{j}=1)\right\|_{2} =  \left\|\frac{1}{|\mathcal{S}_{0}|} \sum_{v_j \in \mathcal{S}_{0}} \mathbf{h}^{l+1}_{j}-\frac{1}{|\mathcal{S}_{1}|} \sum_{v_j \in \mathcal{S}_{1}} \mathbf{h}^{l+1}_{j}\right\|_{2}.
\end{equation}
Let's re-define aggregated representation for node $i$ at $l$th GAT layer as $\bbz^{l+1}_{i}= \sum_{v_j \in \mathcal{N}_{i}} \alpha^{l}_{i j} \bbc^{l+1}_{j}$ for GATs, where $\bbc^{l+1}_{i}=\mathbf{W}^{l} \bbh^{l}_{i}$. 
Lemma \ref{lemma:activated} shows that the deviation between the output representations can be upper bounded by the following term:
\begin{equation}
\label{eq:lem}
    \left\|\frac{1}{|\mathcal{S}_{0}|} \sum_{v_j \in \mathcal{S}_{0}} \mathbf{h}^{l+1}_{j}-\frac{1}{|\mathcal{S}_{1}|} \sum_{v_j \in \mathcal{S}_{1}} \mathbf{h}^{l+1}_{j}\right\|_{2} \leq  L\left( \left\|\frac{1}{|\mathcal{S}_{0}|} \sum_{v_j \in \mathcal{S}_{0}} \mathbf{z}^{l+1}_{j}-\frac{1}{|\mathcal{S}_{1}|} \sum_{v_j \in \mathcal{S}_{1}} \mathbf{z}^{l+1}_{j}\right\|_{2} + 2 \sqrt{N} \Delta^{l+1}_{z} \right),
\end{equation}
where $\Delta^{l+1}_{z}$ is the maximal deviation of aggregated representations $\mathbf{Z}^{l+1}$ at the $l$th GAT layer. Based on this upper bound, the term 
$ \left\|\frac{1}{|\mathcal{S}_{0}|} \sum_{v_j \in \mathcal{S}_{0}} \mathbf{z}^{l+1}_{j}-\frac{1}{|\mathcal{S}_{1}|} \sum_{v_j \in \mathcal{S}_{1}} \mathbf{z}^{l+1}_{j}\right\|_{2}$ will be analyzed. We first consider the terms $\frac{1}{|\mathcal{S}_{1}|}\sum_{v_{j} \in S_{1}} \mathbf{z}^{l+1}_{j}$ and $\frac{1}{|\mathcal{S}_{0}|}\sum_{v_{j} \in S_{0}} \mathbf{z}^{l+1}_{j}$ individually. Based on the assumptions A1 in the main text, the following can be written:
\begin{equation}
\label{eq:zu}
\begin{split}
    \frac{1}{|\mathcal{S}_{1}|}\sum_{v_{j} \in S_{1}} \mathbf{z}^{l+1}_{j} &\in \frac{1}{|\mathcal{S}_{1}|}(\sum_{v_{k} \in S^{\chi}_{1}}(\sum_{a \in \mathcal{N}(k) \cap S_0} \alpha^{l}_{k a} \bar{\mathbf{c}}_{0}^{l+1} + \sum_{b \in \mathcal{N}(k) \cap S_1} \alpha^{l}_{k b} \bar{\mathbf{c}}_{1}^{l+1} )\\
    &+ \sum_{v_{n} \in S^{\omega}_{1}} \sum_{o \in \mathcal{N}(n) \cap S_1} \alpha^{l}_{n o} \bar{\mathbf{c}}_{1}^{l+1}) \pm \Delta^{l+1}_{c}\boldsymbol{1}, \\
    &\in \frac{1}{|\mathcal{S}_{1}|}(\sum_{v_{k} \in S^{\chi}_{1}}(\alpha^{\chi} \bar{\mathbf{c}}_{0}^{l+1} + \alpha^{\omega} \bar{\mathbf{c}}_{1}^{l+1}) + \sum_{v_{n} \in S^{\omega}_{1}}\bar{\mathbf{c}}_{1}^{l+1}) \pm \Delta^{l+1}_{c}\boldsymbol{1} , \\
    &\in \frac{|S^{\chi}_{1}|}{|\mathcal{S}_{1}|} (\alpha^{\chi} \bar{\mathbf{c}}_{0}^{l+1} + \alpha^{\omega} \bar{\mathbf{c}}_{1}^{l+1}) + \frac{|S^{\omega}_{1}|}{|\mathcal{S}_{1}|} \bar{\mathbf{c}}_{1}^{l+1} \pm \Delta^{l+1}_{c}\boldsymbol{1},
    \end{split}
\end{equation}
where $\alpha^{\chi}=\sum_{a \in \mathcal{N}(k) \cap S_0} \alpha^{l}_{k a}$ and $\alpha^{\omega}=\sum_{b \in \mathcal{N}(k) \cap S_1} \alpha^{l}_{k b}$ for a node $v_{k} \in \mathcal{S}_{1}$ based on assumption A2 in the main text with $\alpha^{\chi}+\alpha^{\omega}=1$, and $\boldsymbol{1} \in \mathbb{R}^{F}$ is a vector with all elements being equal to $1$. Let's define $R_{1}^{\chi}:=\frac{|S^{\chi}_{1}|}{|\mathcal{S}_{1}|}$ and $R_{0}^{\chi}:=\frac{|S^{\chi}_{0}|}{|\mathcal{S}_{0}|}$, where $R_{1}^{\omega}= 1- R_{1}^{\chi}$, $R_{0}^{\omega}= 1- R_{0}^{\chi}$.
Similarly, the expression for the term $\frac{1}{|\mathcal{S}_{0}|}\sum_{v_{j} \in S_{0}} \mathbf{z}^{l+1}_{j}$ can also be derived as
\begin{equation}
\label{eq:zv}
    \frac{1}{|\mathcal{S}_{0}|}\sum_{v_{j} \in S_{0}} \mathbf{z}_{j}^{l+1} \in R_{0}^{\chi} (\alpha^{\chi} \bar{\mathbf{c}}_{1}^{l+1} + \alpha^{\omega} \bar{\mathbf{c}}_{0}^{l+1} )+ R_{0}^{\omega} \bar{\mathbf{c}}_{0}^{l+1} \pm \Delta^{l+1}_{c}\boldsymbol{1}.
\end{equation}
Define $\bbepsilon^{l+1}:=\frac{1}{|\mathcal{S}_{1}|}\sum_{v_{j} \in S_{1}} \mathbf{z}^{l+1}_{j} - \frac{1}{|\mathcal{S}_{0}|}\sum_{v_{j} \in S_{0}} \mathbf{z}^{l+1}_{j}$, the following can be written 
\begin{equation}
\label{eq:consider_r}
    \bbepsilon^{l+1} \in \bar{\mathbf{c}}_{0}^{l+1} (R_{1}^{\chi}\alpha^{\chi} - R_{0}^{\chi}\alpha^{\omega} - R_{0}^{\omega}) - \bar{\mathbf{c}}_{1}^{l+1} (R_{0}^{\chi}\alpha^{\chi} - R_{1}^{\chi}\alpha^{\omega} - R_{1}^{\omega}) \pm 2\Delta^{l+1}_{c}\boldsymbol{1}.
\end{equation}
Use $R_{1}^{\omega}= 1- R_{1}^{\chi}$ and $R_{0}^{\omega}= 1- R_{0}^{\chi}$:
\begin{equation}
\label{eq:consider_r1}
    \bbepsilon^{l+1} \in \bar{\mathbf{c}}_{0}^{l+1} (R_{1}^{\chi}\alpha^{\chi} - R_{0}^{\chi}\alpha^{\omega} - 1 + R_{0}^{\chi}) - \bar{\mathbf{c}}_{1}^{l+1} (R_{0}^{\chi}\alpha^{\chi} - R_{1}^{\chi}\alpha^{\omega} - 1 + R_{1}^{\chi}) \pm 2\Delta^{l+1}_{c}\boldsymbol{1}.
\end{equation}
Use $\alpha^{\chi}+\alpha^{\omega}=1$:
\begin{equation}
\label{eq:consider_r2}
\begin{split}
    \bbepsilon^{l+1} &\in  \bar{\mathbf{c}}_{0}^{l+1} (R_{1}^{\chi}\alpha^{\chi} + R_{0}^{\chi}\alpha^{\chi} - 1) -  \bar{\mathbf{c}}_{1}^{l+1} (R_{0}^{\chi}\alpha^{\chi} + R_{1}^{\chi}\alpha^{\chi} - 1 ) \pm 2\Delta^{l+1}_{c}\boldsymbol{1}, \\
    &= (\bar{\mathbf{c}}_{0}^{l+1} - \bar{\mathbf{c}}_{1}^{l+1})((R_{1}^{\chi}\alpha^{\chi} + R_{0}^{\chi}\alpha^{\chi} - 1)) \pm 2\Delta^{l+1}_{c}\boldsymbol{1}.
    \end{split}
\end{equation}
Therefore,  $\|\bbepsilon^{l+1}\|_{2}$ can be upper bounded by:
\begin{equation}
\label{eq:fupper}
   \|\bbepsilon^{l+1}\|_{2} \leq  |(R_{1}^{\chi}\alpha^{\chi} + R_{0}^{\chi}\alpha^{\chi} - 1)| \|\bar{\mathbf{c}}_{0}^{l+1} - \bar{\mathbf{c}}_{1}^{l+1}\|_{2} + 2 \sqrt{N} \Delta^{l+1}_{c} .
\end{equation}
Furthermore, consider the term $\|\bar{\mathbf{c}}_{0}^{l+1} - \bar{\mathbf{c}}_{1}^{l+1}\|_{2}$, where $\bar{\mathbf{c}}_{0}^{l+1} ={\rm mean}(\mathbf{c}^{l+1}_j \mid v_{j} \in \mathcal{S}_{s})$ and $\bbc^{l+1}_{j}=\mathbf{W}^{l} \bbh^{l}_{j}$.
\begin{equation}
\label{eq:sing}
\begin{split}
     \|\frac{1}{|\mathcal{S}_{0}|}\sum_{v_{j} \in S_{0}} \mathbf{c}_{j}^{l+1} - \frac{1}{|\mathcal{S}_{1}|}\sum_{v_{j} \in S_{1}} \mathbf{c}_{j}^{l+1}\|_{2} &= \|\frac{1}{|\mathcal{S}_{0}|}\sum_{v_{j} \in S_{0}} \mathbf{W}^{l} \bbh^{l}_{j} - \frac{1}{|\mathcal{S}_{1}|}\sum_{v_{j} \in S_{1}} \mathbf{W}^{l} \bbh^{l}_{j}\|_{2}\\
     &=\| \mathbf{W}^{l} (\frac{1}{|\mathcal{S}_{0}|}\sum_{v_{j} \in S_{0}} \bbh^{l}_{j} - \frac{1}{|\mathcal{S}_{1}|}\sum_{v_{j} \in S_{1}}  \bbh^{l}_{j})\|_{2}   \\
     &\leq \sigma_{max}(\mathbf{W}^{l}) \|\frac{1}{|\mathcal{S}_{0}|}\sum_{v_{j} \in S_{0}} \bbh^{l}_{j} - \frac{1}{|\mathcal{S}_{1}|}\sum_{v_{j} \in S_{1}}  \bbh^{l}_{j}\|_{2},
     \end{split}
\end{equation}
where $ \sigma_{max}(.)$ outputs the largest singular value of the input matrix. Basen on \eqref{eq:fupper} and \eqref{eq:sing}, it follows that:
\begin{equation}
\label{eq:supper}
    \|\bbepsilon^{l+1}\|_{2} \leq  \sigma_{max}(\mathbf{W}^{l}) |(R_{1}^{\chi}\alpha^{\chi} + R_{0}^{\chi}\alpha^{\chi} - 1)| \|\frac{1}{|\mathcal{S}_{0}|}\sum_{v_{j} \in S_{0}} \bbh^{l}_{j} - \frac{1}{|\mathcal{S}_{1}|}\sum_{v_{j} \in S_{1}}  \bbh^{l}_{j}\|_{2} + 2 \sqrt{N} \Delta^{l+1}_{c} .
\end{equation}
Finally, combining the results in \eqref{eq:lem} and \eqref{eq:supper}, deviation between the output representations from different sensitive groups can be upper bounded by:
\begin{equation}
\begin{split}
&\left\|\frac{1}{|\mathcal{S}_{0}|} \sum_{v_j \in \mathcal{S}_{0}} \mathbf{h}^{l+1}_{j}-\frac{1}{|\mathcal{S}_{1}|} \sum_{v_j \in \mathcal{S}_{1}} \mathbf{h}^{l+1}_{j}\right\|_{2} \leq  L\left( \left\| \bbepsilon^{l+1} \right\|_{2} + 2 \sqrt{N} \Delta^{l+1}_{z} \right),\\
&\leq L\left(  \sigma_{max}(\mathbf{W}^{l}) |(R_{1}^{\chi}\alpha^{\chi} + R_{0}^{\chi}\alpha^{\chi} - 1)| \|\frac{1}{|\mathcal{S}_{0}|}\sum_{v_{j} \in S_{0}} \bbh^{l}_{j} - \frac{1}{|\mathcal{S}_{1}|}\sum_{v_{j} \in S_{1}}  \bbh^{l}_{j}\|_{2} + 2 \sqrt{N} \Delta^{l+1}_{c} + 2 \sqrt{N} \Delta^{l+1}_{z} \right).
\end{split}
\end{equation}

\section{Proof of Lemma \ref{lemma:mlp}}
\label{app:lemma_mlp}
Lemma \ref{lemma:activated} shows that the deviation between the output representations can be upper bounded by the following term:
\begin{equation}
\label{eq:mlp_lem}
\begin{split}
    \delta_{h}^{l+1}:=\left\|\operatorname{mean}(\mathbf{h}^{l+1}_{j} \mid s_{j}=0) - \operatorname{mean}(\mathbf{h}^{l+1}_{j} \mid s_{j}=1)\right\|_{2}& = 
    \left\|\frac{1}{|\mathcal{S}_{0}|} \sum_{v_j \in \mathcal{S}_{0}} \mathbf{h}^{l+1}_{j}-\frac{1}{|\mathcal{S}_{1}|} \sum_{v_j \in \mathcal{S}_{1}} \mathbf{h}^{l+1}_{j}\right\|_{2} \\
    &\leq  L\left( \left\|\frac{1}{|\mathcal{S}_{0}|} \sum_{v_j \in \mathcal{S}_{0}} \mathbf{z}^{l+1}_{j}-\frac{1}{|\mathcal{S}_{1}|} \sum_{v_j \in \mathcal{S}_{1}} \mathbf{z}^{l+1}_{j}\right\|_{2} + 2 \sqrt{N} \Delta^{l+1}_{z} \right),
    \end{split}
\end{equation}
where $\Delta^{l+1}_{z}$ is the maximal deviation of aggregated representations $\mathbf{Z}^{l+1}=\mathbf{H}^{l} \mathbf{W}^{l}$ at the $l$th fully connected layer.
The deviation between aggregated representations from different sensitive groups can be upper bounded by:
\begin{equation}
\label{eq:mlp_upp}
    \begin{split}
     \left\|\frac{1}{|\mathcal{S}_{0}|} \sum_{v_j \in \mathcal{S}_{0}} \mathbf{z}^{l+1}_{j}-\frac{1}{|\mathcal{S}_{1}|} \sum_{v_j \in \mathcal{S}_{1}} \mathbf{z}^{l+1}_{j}\right\|_{2} &= \left\|\frac{1}{|\mathcal{S}_{0}|} \sum_{v_j \in \mathcal{S}_{0}} \mathbf{W}^{l} \mathbf{h}^{l}_{j}-\frac{1}{|\mathcal{S}_{1}|} \sum_{v_j \in \mathcal{S}_{1}} \mathbf{W}^{l} \mathbf{h}^{l}_{j}\right\|_{2} \\
     &= \left\| \mathbf{W}^{l} (\frac{1}{|\mathcal{S}_{0}|} \sum_{v_j \in \mathcal{S}_{0}} \mathbf{h}^{l}_{j}-\frac{1}{|\mathcal{S}_{1}|} \sum_{v_j \in \mathcal{S}_{1}}\mathbf{h}^{l}_{j})\right\|_{2}\\
     & \leq \sigma_{max}(\mathbf{W}^{l}) \left\|(\frac{1}{|\mathcal{S}_{0}|} \sum_{v_j \in \mathcal{S}_{0}} \mathbf{h}^{l}_{j}-\frac{1}{|\mathcal{S}_{1}|} \sum_{v_j \in \mathcal{S}_{1}}\mathbf{h}^{l}_{j})\right\|_{2}.
    \end{split}
\end{equation}
Therefore, combining the results of \eqref{eq:mlp_lem}, \eqref{eq:mlp_upp}, it follows that:
\begin{equation}
     \left\|\frac{1}{|\mathcal{S}_{0}|} \sum_{v_j \in \mathcal{S}_{0}} \mathbf{h}^{l+1}_{j}-\frac{1}{|\mathcal{S}_{1}|} \sum_{v_j \in \mathcal{S}_{1}} \mathbf{h}^{l+1}_{j}\right\|_{2} \leq  L \left(  \sigma_{max}(\mathbf{W}^{l}) \left\|(\frac{1}{|\mathcal{S}_{0}|} \sum_{v_j \in \mathcal{S}_{0}} \mathbf{h}^{l}_{j}-\frac{1}{|\mathcal{S}_{1}|} \sum_{v_j \in \mathcal{S}_{1}}\mathbf{h}^{l}_{j})\right\|_{2} + 2 \sqrt{N} \Delta^{l+1}_{z}\right),
\end{equation}
which concludes the proof.

\section{Additional Statistics on Datasets}
\label{app:data_stats}

Further statistical information for the utilized datasets are presented in Tables \ref{table:stats} and \ref{table:stats_non_binary}.  $F$ in Table \ref{table:stats} denotes the dimension of nodal features. 
\begin{table}[ht]
	\centering
\caption{Dataset statistics for social networks}
\label{table:stats}
\begin{footnotesize}
\begin{tabular}{c c c c c c c}
\toprule
                                                    Dataset&  $|\mathcal{S}_{0}|$ & $|\mathcal{S}_{1}|$ & $| \mathcal{E}^{\chi}|$& $| \mathcal{E}^{\omega}|$ & $F$                   \\ 
\midrule
Pokec-z  & $4851$ & $2808$ & $1140$ & $28336$ & $59$ \\
Pokec-n  & $4040$ & $2145$ & $943$ & $20901$ & $59$ \\
Recidivism  & $9317$ & $9559$ & $298098$ & $325642$ & $17$ \\
  \bottomrule
\end{tabular}
\end{footnotesize}
\end{table}

\begin{table}[h!]
	\centering
\caption{Dataset statistics for citation networks}
\label{table:stats_non_binary}
\begin{footnotesize}
\begin{tabular}{c c c c c}
\toprule
                                                    Dataset&  $|\mathcal{V}|$ &
                                                    \# sensitive attr. &
                                                    $|\mathcal{E}^{\chi}|$ & $|\mathcal{E}^{\omega}|$                       \\ 
\midrule
Cora & $2708$ & $7$ & $1428$ & $5964$ \\
Citeseer & $3327$ & $6$ & $1628$ & $4746$  \\
Pubmed & $19717$ & $3$ & $12254$ & $49802$ 
 \\ \bottomrule
\end{tabular}
\end{footnotesize}
\end{table}

\section{Hyperparameters}
\label{app:hypers}
We provide the selected hyperparameter values for the GNN model and the proposed framework, FairGAT, for the reproducibility of the presented results. For the node classification task, weights are initialized utilizing Glorot initialization \cite{glorot} in the GAT-based classifier. All models are trained for $500$ epochs by employing Adam optimizer \cite{adam} together with a learning rate of $0.005$ and $\ell_2$ weight decay factor of $0.0005$. A $2$-layer GAT network followed by a linear layer is employed for node classification (for baselines the conventional GAT layer \cite{gat} is utilized, while the layer used in FairGAT employs fair attention calculation described in \eqref{eq:fairatt} ). Hidden dimension of the node representations is selected as $128$ on all datasets. The experimental settings for link prediction are kept the same as in FairDrop \cite{spinelli2021fairdrop}, where GAT layers are utilized to build the encoder network. 

The results for baseline schemes, FairGNN \cite{say}, EDITS \cite{dong2022edits}, and NIFTY \cite{nifty} are obtained by choosing the hyperparameters in the corresponding studies via grid search on cross-validation sets with $5$ different data splits. Specifically, the values $0.01, 0.1, 0.2, 1.$ are examined as the multiplier of adversarial regularizer for FairGNN. The results in Table \ref{table:comp} are obtained by setting the value of this hyperparameter to $0.01, 0.2$ and $0.1$ for Pokec-z, Pokec-n and Recidivism datasets, respectively.  Moreover, the threshold values $ 0.001, 0.01, 0.05, 0.3$ are examined for EDITS (except for Recidivism dataset, for which the study already suggests the optimal hyperparameter value $0.015$). The results in Table \ref{table:comp} are obtained for $0.001$ and $0.05$ for Pokec-z and Pokec-n, respectively. Finally for NIFTY, the coefficient of the unsupervised loss is tuned among the values  $0.5, 0.6, 0.7, 0.8, 0.9$. The results in Table \ref{table:comp} are obtained for $0.9, 0.5$ and $0.5$ for Pokec-z, Pokec-n, and Recidivism, respectively. 

Hyperparameter values for $\alpha_{max}^{\chi}$ and $\eta$ that lead to the results in Tables \ref{table:comp} and \ref{table:fairlp} are presented in Table \ref{table:method_params}. In this work, these hyperparameters are selected via grid search on cross-validation sets over $5$ different data splits. Specifically, the values $0.25, 0.50, 0.75$ are examined for $\alpha_{max}^{\chi}$. In the tuning of $\eta$, for a more stable tuning process, both $\mathbf{Z}^{l+1}$ and  $\mathbf{C}^{l+1}$ are normalized, so that the variation of each feature equals to $1$, before scaling with $\eta$. This normalization before scaling also allows the overall scaling factors to be different for different layers $l$, which provides a better fairness-utility trade-off for different $\eta$ values. The values $0.75, 1$ are examined for the final $\eta$ selection.
\begin{table}[ht]
	\centering
\caption{Utilized $\alpha_{max}^{\chi}$ and $\eta$ values for the presented results in Tables \ref{table:comp} and \ref{table:fairlp}}
\label{table:method_params}
\begin{tabular}{c c c c c c c}
\toprule
                                           &        Pokec-z&  Pokec-n&  Recidivism &  Cora &  Citeseer &  Pubmed     \\ 
\midrule
$\alpha_{max}^{\chi}$  & $0.75$ & $0.25$ & $0.75$ & $0.50$ & $0.75$ & $0.50$ \\
$\eta$  & $1.00$ & $0.75$ & $1.00 $ & $1.00 $ & $1.00 $ & $1.00 $\\
 \bottomrule
\end{tabular}
\end{table}

\section{Computing Infrastructures}
\label{app:computing}
\textbf{Software infrastructures:} All GNN-based models are trained utilizing PyTorch 1.10.1 \cite{pytorch} and NetworkX 2.5.1 \cite{networkx}.

\textbf{Hardware infrastructures:} Experiments are carried over on 8 AMD Ryzen Threadripper 3970X CPUs.

\end{document}